\documentclass[journal]{IEEEtran}
\usepackage{amsmath,amsfonts}
\usepackage{algorithm}
\usepackage{array}
\usepackage[caption=false,font=normalsize,labelfont=sf,textfont=sf]{subfig}
\usepackage{textcomp}
\usepackage{stfloats}
\usepackage{url}
\usepackage{verbatim}
\usepackage{graphicx}
\usepackage{cite}
\usepackage{booktabs} 
\usepackage[noend]{algpseudocode}
\usepackage{xcolor}
\usepackage{lipsum}
\usepackage{enumitem}
\usepackage{bm}
\usepackage{makecell}
\usepackage{amssymb} 
\newcommand{\replications}[0]{10}
\newcommand{\replicationsparams}[0]{5}
\newcommand{\replicationsreeval}[0]{10}
\newcommand{\totalreplications}[0]{1120}
\newcommand{\totalreplicationsparams}[0]{1520}

\newcommand{\samplingsize}[0]{sampling-size}
\newcommand{\archivesampling}[0]{Archive-sampling}
\newcommand{\eas}[0]{Parallel-Adaptive-sampling}
\newcommand{\uqdlong}[0]{Uncertain Quality-Diversity}
\newcommand{\uqd}[0]{Uncertain QD}
\newcommand{\code}[0]{\url{https://github.com/adaptive-intelligent-robotics/Uncertain_Quality_Diversity}}

\begin{document}

\title{


\uqdlong{}: \\
Evaluation methodology and new methods for Quality-Diversity in Uncertain Domains




}

\author{Manon Flageat and Antoine Cully, ~\IEEEmembership{Adaptive and Intelligent Robotics Lab, Imperial College London}
}



\maketitle

\begin{abstract}
Quality-Diversity optimisation (QD) has proven to yield promising results across a broad set of applications. 
However, QD approaches struggle in the presence of uncertainty in the environment, as it impacts their ability to quantify the true performance and novelty of solutions.
This problem has been highlighted multiple times independently in previous literature. 
In this work, we propose to uniformise the view on this problem through four main contributions. 
First, we formalise a common framework for uncertain domains: the \uqd{} setting, a special case of QD in which fitness and descriptors for each solution are no longer fixed values but distribution over possible values.
Second, we propose a new methodology to evaluate \uqd{} approaches, relying on a new per-generation sampling budget and a set of existing and new metrics specifically designed for \uqd{}.
Third, we propose three new \uqd{} algorithms: \archivesampling{}, \eas{} and Deep-Grid-sampling. 
We propose these approaches taking into account recent advances in the QD community toward the use of hardware acceleration that enable large numbers of parallel evaluations and make sampling an affordable approach to uncertainty.
Our final and fourth contribution is to use this new framework and the associated comparison methods to benchmark existing and novel approaches.
We demonstrate once again the limitation of MAP-Elites in uncertain domains and highlight the performance of the existing Deep-Grid approach, and of our new algorithms. 
The goal of this framework and methods is to become an instrumental benchmark for future works considering \uqd{}. 

\end{abstract}


\section{Introduction}

Recent works have shown the competitiveness of diversity-seeking approaches to optimisation, such as Quality-Diversity (QD) \cite{framework, pugh, book_chapter}. 
Encouraging creative and diverse solutions have proven to help exploration \cite{design, video_games}, foster stepping stones toward novel solutions \cite{stepping_stones} and allow adaptation to unseen situations \cite{nature}. 
QD in particular has been applied to a wide range of domains ranging from robotics \cite{robotic, nature} to content generation \cite{video_games, video_games_matt} or design \cite{design}.
Recently, many QD works have shifted focus toward Neuroevolution and the evolution of large closed-loop controllers for robotics \cite{benchmark, pga, mapes, qdpg}.

\begin{figure}[t!]
\centering
\includegraphics[width = \hsize]{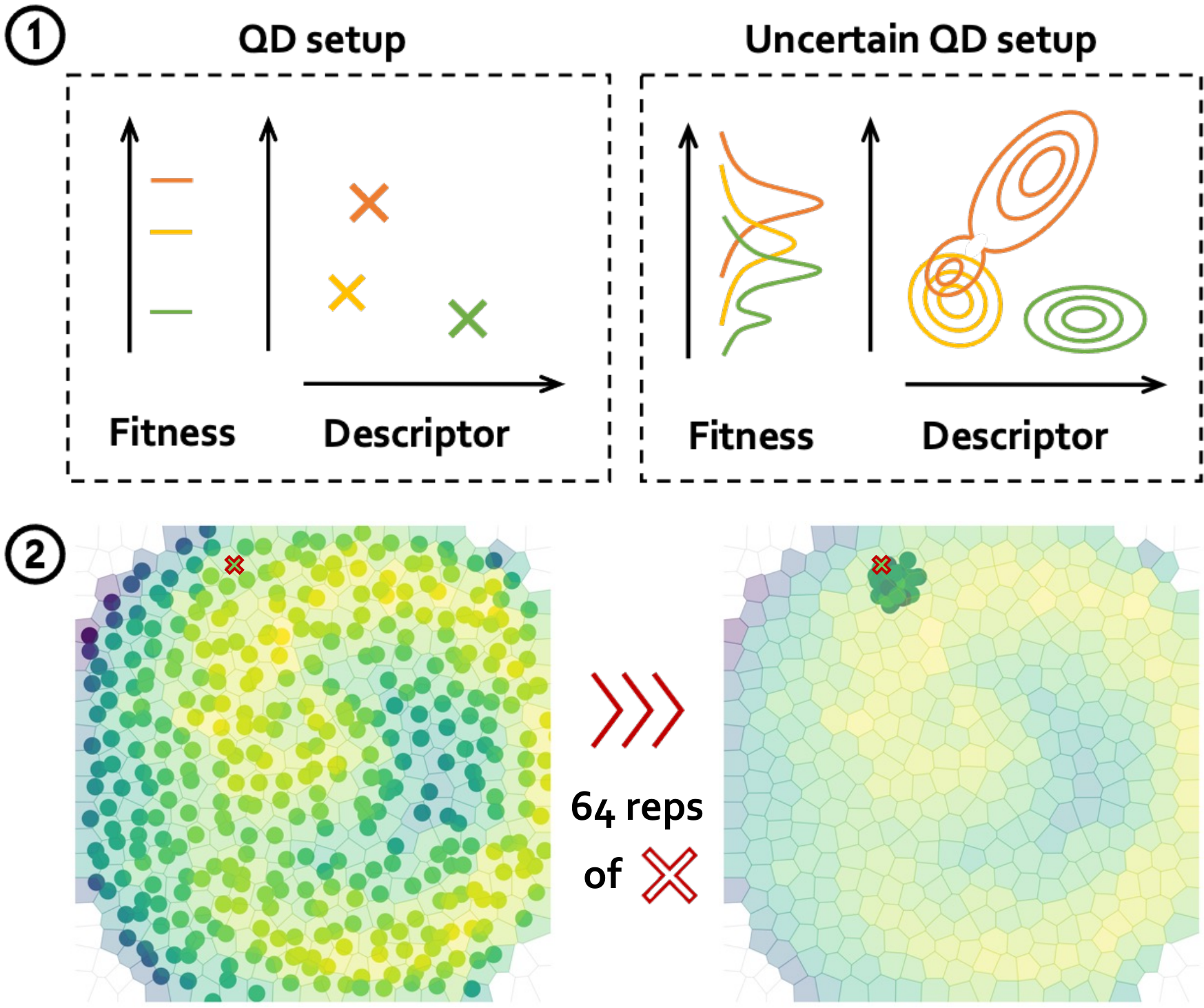}
\caption{
Illustration of the \uqd{} setting:
(1) we represent the fitness and descriptors of $3$ hypothetical solutions in the usual QD setup (left) versus in the \uqd{} setup (right): their fitness and descriptors are now distributions over possible values, instead of fixed constant values.
(2) We give the final archive of the Uncertain Arm task (left), we sample one of the solution (red cross), replicate it $64$ times and plot the descriptor and fitness distributions, keeping the archive in the background for comparability (right). 
}
\label{fig:uncertain}
\end{figure}

These new domains raise new challenges for QD optimisation, among which the issue of uncertainty \cite{adaptive, deepgrid, glette_stochastic, pga_stochastic}.
Many environments are subject to uncertainty, either arising in components such as actuators or sensors, or occurring in the dynamics of the environment itself. 
This uncertainty might lead the exact same solution to perform differently from one evaluation to another. 
Thus, evaluating a solution only once might lead to an inaccurate estimate of its novelty or quality: solutions might be lucky and appear more novel or better performing than they truly are. 
We illustrate this issue in Fig.~\ref{fig:uncertain}.
Unfortunately, standard QD algorithms maintain lucky solutions due to their elitism, leading them to prefer lucky solutions to truly novel or high-performing ones \cite{adaptive, deepgrid, glette_stochastic, pga_stochastic}. 
While this issue is well known in the larger Evolutionary Algorithm (EA) community \cite{ea_uncertain, ea_uncertain_2}, to the best of our knowledge, it stays largely understudied in QD works. 

A naive approach to address this uncertainty problem is to perform an extensive and expensive amount of evaluations for each solution to obtain the average or expected performance and novelty metrics. 
While this approach is effective it is often prohibitively computationally expensive. Nonetheless, recent advances in computer systems enable the high-parallelisation of evaluations.
Recent libraries such as QDax \cite{qdax}, or EvoJax \cite{evojax} based on the Brax simulator \cite{brax} allowed to speed-up computation by a large order of magnitude thanks to the high-parallelisation of evaluations.
With such tools, we now have access to $10$ or $100$ times more evaluations per generation within the same amount of time. 
This completely changes the game for \uqd{}, as simpler sample-heavy approaches could now be more interesting than more advanced model-driven approaches.
However, this shift toward sample-heavy approaches to uncertainty should also lead to reconsider how algorithms are compared.
Most QD works use comparison based on batch-size, i.e. number of offspring produced per generation.  
However, if every offspring require multiple evaluations, these additional samples should also be taken into account. 
We propose to replace batch-size with sampling-size: per-generation sampling-budget allowing fair comparison of strategies to uncertainty.

Overall, our paper aims to answer one core question: in an uncertain domain, given a per-generation sampling budget, what would be the optimal \uqd{} approach? Our contribution can be summarised as follows:
\begin{itemize}
    \item A formalisation of the \uqd{} setting.
    \item A new methodology to evaluate \uqd{} approaches relying on the sampling-size introduced above, as well as a set of new and existing metrics.
    \item Three new \uqd{} approaches: \archivesampling{}, \eas{} and Deep-grid-sampling.
    \item A comparison of existing and novel \uqd{} approaches using these new comparison methods.
\end{itemize}
To facilitate later advances in \uqd{} we open-source our code at \code{}.

\section{Background}

\subsection{Quality-Diversity and MAP-Elites algorithm}

Quality-Diversity (QD) methods propose to search for collections of high-performing and diverse solutions to an optimisation problem. 
In the QD framework, a solution does not only get attributed a fitness $f$ but also a descriptor $d$. The descriptor $d$ characterises the behaviour induced by the solution, according to a fixed number $n$ of dimensions of interest $d = (d^j)_{1 \leq j \leq n}$. It allows one to quantify the novelty of a solution, as the  average distance to its nearest neighbours.

The most widely used QD algorithm is MAP-Elites \cite{map_elites}, the focus of this work.
MAP-Elites discretises the descriptor-space into a grid, referred to as archive $\mathcal{A}$. The goal of MAP-Elites is to fill all the cells of this grid with the best possible solutions. This grid constitutes the final collection returned at the end of the algorithm. 
The archive $\mathcal{A}$ is first initialised with randomly-generated solutions. Then, an iteration of MAP-Elites consists of the following steps: 1. sample uniformly parent solutions from the archive $\mathcal{A}$, 2. apply variation operators to these solutions to produce offspring solutions, 3. evaluate the offspring solutions to get their descriptor $d$ and fitness $f$, 
and finally 4. attempt to add the offspring back to the archive $\mathcal{A}$. For a new solution to be added to the grid, it has to improve either the quality or the diversity of the overall grid. Thus, it has to either fill in an empty cell or perform better than the solution already in its cell.
Overall, MAP-Elites aims to maximise the QD-Score of the archive $\mathcal{A}$, defined as follow:
\begin{equation} \label{eq:qd_obj}
\begin{split}
    \max_{\mathcal{A}} \left( \textrm{QDScore} (\mathcal{A}) \right) = \max_{\mathcal{A}} \sum_{i \in \mathcal{A}}{f_i} \\
    \text{w.r.t} \quad \forall i \in \mathcal{A}, d_i\in \textrm{cell}_i
\end{split}
\end{equation}
Where $f_i$ is the fitness of solution $i$ and $d_i$ is its descriptor, which determines its cells in the archive $cell_i$.

\subsection{Hardware-accelerated Quality-Diversity}

Recent advances in hardware acceleration have led to new QD libraries such as QDax \cite{qdax} or EvoJax \cite{evojax}. 
These tools rely on highly-parallelised simulators like Brax \cite{brax} that can run on accelerators (e.g., GPUs and TPUs) and thus target simulated domains, for example, robotics control, where they drastically reduce the evaluation time. 
In addition, they have given us access to $10$ or $100$ times more evaluations per generation within the same amount of time.
Lim et al. \cite{qdax} prove that the performance of MAP-Elites is robust to large increases in batch-size values (i.e. large increases in the number of solutions generated per generation). These results lead to drastic speed ups in the run-time of QD algorithms, opening the door to promising future applications.
These new frameworks completely change the game for \uqd{}, as naive approaches using extensive and expensive amount of evaluations for each solution  could now be more interesting than more advanced model-driven approaches.

\section{\uqd{} setting} \label{sec:uncertain}

\subsection{Uncertain domains}

Usually, QD algorithms rely on the hypothesis that each solution can be awarded a fixed fitness $f$ and descriptor $d$, i.e. the reevaluation of a solution would give the same results as its initial evaluation. 
In this work, we consider domains where this assumption does not hold: a solution can be assigned different values of fitness and descriptor from one evaluation to another.
Examples of such domains would be real-world applications where the sensors used to compute the fitness and the descriptor of a solution are slightly noisy or inaccurate. Similarly, in locomotion robotics, the initialisation of the robot might change from one evaluation to another, leading to different behaviours for the same controller.
In other words, each solution is not assigned one value of fitness and descriptor anymore, but a distribution of potential fitness and descriptor values: $f \sim \mathcal{D}_f$, $d \sim \mathcal{D}_d$.
Such domains have been extensively studied in the Evolutionary Algorithm literature where only the fitness is defined \cite{ea_uncertain, ea_uncertain_2}, and referred to as uncertain domains. 
Inspired by these works, we propose the \uqd{} setting, illustrated in Figure \ref{fig:uncertain}.
We would like to emphasise that \uqd{} with a null variance on both fitness and descriptor corresponds to the usual QD case.

\subsection{Impact of uncertainty on QD approaches}
In uncertain domains, evaluating a solution only once might lead to inaccurate estimates of its diversity or quality: this unique evaluation might be an outlier and not be representative of the fitness and descriptor distributions as a whole. 
This outlier evaluation might appear more novel or higher performing than the solution actually is when averaging multiple evaluations.
As QD algorithms are elitists, they tend to maintain solutions that get such lucky outlier evaluations. 
This leads to final collections containing a majority of lucky sub-performing solutions instead of truly novel and high-performing ones \cite{deepgrid, adaptive, pga_stochastic, glette_stochastic}. 
Instead, QD approaches designed for uncertain domains aim to get high certitude on the descriptor and fitness estimate that can be consistently observed across multiple evaluations of the same solution. 
Thus, Eq.~\ref{eq:qd_obj} becomes:
\begin{equation} \label{eq:qd_uncertain_obj}
\begin{split}
    \max_{\mathcal{A}} \left( \textrm{QDScore} (\mathcal{A}) \right) = \max_{\mathcal{A}} \sum_{i \in \mathcal{A}}{\mbox{E}_{f_i \sim \mathcal{D}_f} \left[ f_i \right]} \\
    \text{w.r.t} \quad \forall i \in \mathcal{A}, \mbox{E}_{d_i \sim \mathcal{D}_d} \left[ d_i \right] \in \textrm{cell}_i
\end{split}
\end{equation}

\subsection{Performance estimation and reproducibility} \label{sec:reproducibility}

To mitigate the impact listed above, two problems should be addressed in uncertain domains:
\begin{itemize}
    \item \textbf{Performance estimation:} getting a good estimate of the expected fitness $\mbox{E}_{f_i \sim \mathcal{D}_f} \left[ f_i \right]$ and expected descriptor $\mbox{E}_{d_i \sim \mathcal{D}_d} \left[ d_i \right]$ of solutions. \uqd{} algorithms need specific mechanisms to estimate the fitness and descriptor distributions and infer meaningful expectations. 
    \item \textbf{Reproducibility:} estimating for each solution the variance of the fitness $\mbox{V}_{f_i \sim \mathcal{D}_f} \left[ f_i \right]$ and the variance of each descriptor dimension $\mbox{V}_{d^j_i \sim \mathcal{D}_{d^j}} \left[ d^j_i \right]$. In uncertain domains, solutions with lower variance are more reproducible: they are more likely to get similar fitness and descriptor values from one evaluation to another. Reproducibility is a desirable property and algorithms need specific mechanisms to first, estimate solutions' reproducibility, and second, to eventually favour it.
\end{itemize}
Not all uncertain domains present the reproducibility problem: all solutions might have the same variance. 
This depends on the properties of the source of uncertainty. 
For example, if the only uncertainty is coming from the sensor giving the fitness that present Gaussian noise on its measurements, all solutions would be identically impacted and there would be no way to improve the reproducibility of solutions.
On the contrary, if the uncertainty is on the initial joint position of a locomotion robot for example, not all controllers would be equally impacted: hopping controllers have higher chances to lead to a fall if the wrong leg is initialised at a high position than controllers using the two legs equally.
Thus, most realistic scenarios suffer from this problem. 
Still, most \uqd{} works consider domains with only the performance estimation problems.

\subsection{Evaluation metrics in \uqd{} setting} \label{sec:uncertain_metrics}

Assessing the performance of QD algorithms usually relies on computing metrics such as QD-Score or Max-Fitness using fitness and descriptors values collected during training.
In \uqd{}, these training data cannot be trusted as they might be poor-quality estimates. 
Also, in most uncertain domains, the ground-truth values are not accessible, even to an expert. 
Thus, in the \uqd{} setting, the evaluation metrics have to be approximated with large number (hundreds) of samples. 
These samples do not count toward the sample efficiency of algorithms, as they are used for metric computation only.

Previous work in \uqd{} \cite{deepgrid, adaptive, pga_stochastic, glette_stochastic} uses as estimations the average fitness and descriptor of $M = 50$ reevaluations of each cell in the algorithm's archive. These estimations are then used to fill a new archive called the "Corrected Archive" that is used to compute metrics such as Coverage and QD-Score. 
For comparability, the $M$ reevaluations of each cell are done by sampling each cell $M$ times using its in-cell selector.
In other words, for approaches that only return the best solution of each cell, the estimation is the average of $M$ replications of the best solutions of each cell; and for approaches considering multiple solutions per-cell, the estimation is the average of $M$ solutions sampled from the cell.

In this work, we also consider a Corrected Archive. However, relying on results from Doerr and Sutton in \cite{median} as well as an experimental study provided in Appendix B, we choose to use the median over $M$ replications instead of the average. 
We also performed a study to choose an appropriate value of $M$ and decided to keep $M=512$ (Appendix B). 

\begin{table*}[ht]
  \caption{Overview of existing and proposed methods designed for \uqd{} setting.}
  \label{tab:baselines}
  \begin{tabular}{ c | c c | c c | c c || c }

    & \multicolumn{2}{c}{\textsc{Type of approach}} & \multicolumn{2}{c}{\textsc{Problem tackled}} & \multicolumn{2}{c}{\textsc{Implementation}} & \\
    
    & \textsc{Explicit} & \textsc{Implicit} & \makecell{\textsc{Perf. estimation}} & \textsc{Reproducibility} & \textsc{Sequential} & \textsc{Parallel} & \textsc{New} \\
    
    \midrule
    
    \textsc{MAP-Elites-sampling} 
    & \checkmark
    & 
    & \checkmark
    &
    &
    & \checkmark  
    &  \\
    
    \textsc{Adaptive-sampling} 
    & \checkmark
    & 
    & \checkmark
    & \checkmark
    & \checkmark
    &  
    &  \\

    \textsc{\archivesampling{}} 
    & \checkmark
    & 
    & \checkmark
    & \checkmark
    &
    & \checkmark  
    & \checkmark  \\

    \textsc{\eas{}} 
    & \checkmark
    & 
    & \checkmark
    & \checkmark
    &
    & \checkmark  
    & \checkmark \\

    \textsc{Deep-Grid} 
    & 
    & \checkmark
    & \checkmark
    & \checkmark
    & 
    & \checkmark  
    & \\

    \textsc{Deep-Grid-sampling} 
    &
    & \checkmark
    & \checkmark
    & \checkmark
    &
    & \checkmark  
    & \checkmark \\

  \end{tabular}
\end{table*}

\section{Previous work in \uqd{}} \label{sec:qd_approaches}

\subsection{Explicit-sampling and Implicit-sampling approaches in EA} \label{sec:explicit_implicit}

The literature in EA applied to uncertain domains distinguishes two main classes of algorithms: explicit sampling and implicit sampling approaches \cite{ea_uncertain, ea_uncertain_2}. 
Explicit-sampling approaches evaluate each solution multiple times to get a better estimate of their fitness distributions \cite{ea_explicit}. The number of evaluations might be fixed or chosen dynamically depending on the methods \cite{ea_adaptive, ea_adaptive_2}.
Implicit-sampling approaches rely on the common hypothesis that the fitness is continuous with respect to the search-space. Thanks to this property, neighbouring solutions can be used as samples to build distribution estimates, thus avoiding re-evaluations.
Most of the time Implicit-sampling approaches consist in augmenting the size of the population -- dynamically or not -- to increase the density of solutions and get better distribution estimates \cite{ea_implicit}. Some Implicit-sampling approaches also propose keeping archives of previously encountered solutions as estimates \cite{ea_archive}.

\subsection{Previous QD approaches}

We apply the same classification to QD approaches, distinguishing explicit and implicit mechanisms to approximate the distribution of fitness and descriptor.
To the best of our knowledge, three QD approaches based on MAP-Elites specifically target uncertain domains: two can be classified as Explicit samplings and one as Implicit sampling. We summarise the properties of these approaches as well as the new ones introduced in this paper in Table~\ref{tab:baselines}.

\subsubsection{MAP-Elites-sampling} \label{sec:naive}

The simplest way to use sampling in MAP-Elites is to set a fixed number of reevaluation $N$ for every solution. Thus, any new offspring $o$ is sampled $N$ times and considered for addition to the grid using the average fitness $\bar{f}_o = \frac{1}{N} \sum_{n=0}^N{f_o^n}$ and descriptor $\bar{d}_o = \frac{1}{N} \sum_{n=0}^N{d_o^n}$.
This approach has been used in previous works \cite{hbr} and proven to greatly improve the performance of MAP-Elites in uncertain domains. 
However, it has three main disadvantages: first, it is highly sample-inefficient, second, it relies on a hyperparameter $N$ that is critical to guarantee a meaningful estimate and third, it improves the performance estimation but does not favour reproducible solutions (see Sec.~\ref{sec:reproducibility}).

\subsubsection{Adaptive sampling} \label{sec:adaptive}
To mitigate MAP-Elites-sampling's limitations, Justesen et al. \cite{adaptive} introduced Adaptive-Sampling. It uses new mechanisms to distribute samples more wisely across solutions, inspired by similar work in EAs \cite{ea_adaptive}:
\begin{itemize}
    \item First, solutions are considered for addition to the grid only if they are evaluated the same number of times as the elite already in their cell. If, before reaching this number of evaluations, they prove not promising enough then they are discarded. This improves sample efficiency. 
    \item Second, if an elite in the grid is challenged by a new solution, it is immediately re-evaluated one more time, allowing to dynamically tune the number of evaluations. If during this re-evaluation, the elite proves more likely to belong to another cell, it is moved to this new cell.
    \item Third, if an elite is evaluated too many times outside its main cell, it is discarded. This mechanism keeps solutions that prove reproducible in terms of descriptor (Sec.~\ref{sec:reproducibility}).
    \item Fourth, to compensate for the loss of elites induced by the two last mechanisms, each cell keeps $D$ elites. Thus, if one of the elites drifts to another cell or is discarded, it can be immediately replaced. 
\end{itemize}
Adaptive-Sampling tackles both the performance estimation and the reproducibility problem, despite being mainly studied in domains where all solutions are equally reproducible. 
However, its main limitation is that it is sequential and thus, not parallelisable.
First, it requires evaluating each offspring the same number of times as the elite of its cell. This number differs between offspring and is likely to change during the process as successive reevaluations might move the offspring to other cells.
Second, it also requires reevaluating contested elites, modifying the repertoire during the reevaluation process.
Thus, these mechanisms make Adaptive-sampling extremely slow in real-time computation, regardless of its sample efficiency.

\subsubsection{Deep-Grid} \label{sec:deepgrid}
 Deep-Grid MAP-Elites \cite{deepgrid} is an alternative Implicit sampling approach to \uqd{}. 
It works by adding a depth to the MAP-Elites grid to keep $D$ solutions per cell, so that each cell sub-population empirically estimates the distribution of the elite's fitness and descriptor. 
To this aim, Deep-Grid modify the grid-maintenance rules:
\begin{itemize}
    \item Each cell contains $D$ solutions, but the best-performing of these $D$ solutions might have been lucky. Thus, the selection is modified to consider all depth-solutions as potential parents: first, a cell is selected uniformly, as in MAP-Elites; second, a solution is selected from the cell using the biased roulette wheel (fitness proportional).
    \item All $D$ solutions in a cell are also assumed to be equally uncertain and should all be questionable, i.e. replaceable. While in MAP-Elites, a high-performing solution can only be replaced by an even higher-performing one; in Deep-Grid, it can be replaced by any solution. Thus, Deep-Grid adds all solutions to the grid. The convergence within the cell relies on the convergence of the population as a whole induced by the MAP-Elites loop.
\end{itemize}
These combined mechanisms allow the cells to be slowly populated with reproductions of reproducible solutions. 
This approach thus tackles both the performance estimation and the reproducibility problems. Its main limitation is its lack of elitism: due to its questioning mechanisms, Deep-Grid struggles to converge to the best-performing solutions.

\subsubsection{Notes on gradient-augmented QD}
Flageat et al.~\cite{pga_stochastic} have shown that gradient-augmented QD approaches such as MAP-Elites-ES~\cite{mapes} or PGA-MAP-Elites~\cite{pga} are promising in the \uqd{} setting. 
However, these gradient-augmented QD approaches do not aim to tackle specifically the \uqd{} setting.
Their performance in these domains only arises as a side effect of their strategy of modelling the environment. 
More importantly, approaches inspired by Reinforcement Learning, such as PGA-MAP-Elites, require additional assumptions on the domains as they rely on Markov Decision Processes (MDPs); and approaches based on approximated gradients such as MAP-Elites-ES require a really large order of evaluations to generate each offspring (MAP-Elites-ES uses $10030$ evaluations per offspring). 
In this paper, we consider both MDP and non-MDP domains, as MDP tasks are only a small portion of the domains considered by QD. 
Additionally, we consider sample efficiency as a critical comparison criterion, and MAP-Elites-ES require orders of magnitude more evaluations than the approaches compared in this paper.
Thus, we chose to exclude these approaches from this study.
As highlighted by Flageat et al.~\cite{pga_stochastic} these approaches stay promising for domains with large-dimensional search space where mutation-based QD approaches prove inefficient. We keep this research dimension for future work.

\section{New approaches for \uqd{} in highly-parallelised context} \label{sec:new_approaches}

Hardware-accelerated libraries require specific adaptation of existing algorithms to leverage their full parallelisation potential. 
Below we detail how MAP-Elites-sampling and Deep-Grid can be adapted to hardware-accelerated libraries. Furthermore, we introduce three novel approaches: Deep-Grid-sampling, \archivesampling{} and \eas{}.
We also detail these algorithms in Table~\ref{tab:baselines}.

\subsection{MAP-Elites-sampling}

Using explicit sampling with MAP-Elites (Sec.~\ref{sec:naive}) does not require any adaptation to high-parallelisation: the samples required to evaluate any solution are fully independent and can be executed in parallel. 
Hardware-accelerated libraries make MAP-Elites-sampling a strong \uqd{} competitor.

\subsection{Deep-Grid}

Deep-Grid (Sec.~\ref{sec:deepgrid}) mainly differs from MAP-Elites in that it adds depth to the grid. The addition and selection from this deep grid remain independent for each solution, making it fully parallelisable. 
However, the Deep-Grid addition mechanism, which adds all offspring to the archive, might be less efficient with a very large batch-size.
First, if the batch-size is greater than the number of solutions in the archive (i.e. grid size times depth), a random subset of the offspring will systematically not be added to the grid, regardless of their fitness and descriptor, making Deep-Grid less sample-efficient.
Second, Deep-Grid relies on the intuition that reproducible solutions slowly populate cells. Large batch-size might lead to renew the full cell content of the grid at every generation, and may limit the effect of this mechanism.

\subsection{Deep-Grid-sampling (novel)} \label{sec:deepgridsampling}

As Deep-Grid's mechanisms do not scale well to large batch-size value, it seems that the available evaluations offered by hardware acceleration could be better employed.
We thus propose Deep-Grid-sampling, a new algorithm that combines Deep-Grid with explicit sampling. 
In Deep-grid-sampling, every offspring is evaluated $N$ times to obtain a more accurate estimation of the descriptor $d$ and the fitness $f$, which are then used to place the offspring in the corresponding cell.

\begin{algorithm}[t!]
    \centering
    \caption{\archivesampling{} (\ref{sec:archivesampling})}\label{algo:archive_sampling}
    \begin{algorithmic}[1]
        \State \text{Initialise $\mathcal{A}$}
        \Repeat
             \vspace{2mm}
            \State \text{\textit{\# Archive reevaluation}}
            \State \text{$x \gets$ content of $\mathcal{A}$}
            \State \text{$\mathcal{A} \gets$ empty($\mathcal{A}$)}
            \State \text{$f$, $d$ $\gets$ reevaluations($x$, $N = 1$)}
            \State \text{$\mathcal{A} \gets$ add($\mathcal{A}$, $x$, $f$, $d$)}
            \vspace{2mm}
            \State \text{\textit{\# Offspring generation}}
            \State \text{$o \gets$ generate\_offspring($\mathcal{A}$)}
            \State \text{$f$, $d$ $\gets$ evaluations($o$, $N = 1$)}
            \State \text{$\mathcal{A} \gets$ add($\mathcal{A}$, $o$, $f$, $d$)}
            \vspace{2mm}
        \Until {reach convergence of $\mathcal{A}$}
    \end{algorithmic}
\end{algorithm}

\subsection{\archivesampling{} (novel)} \label{sec:archivesampling}

We propose a new approach for uncertain domain: \archivesampling{} (Alg.~\ref{algo:archive_sampling}).
\archivesampling{} aims to improve MAP-Elites-sampling by distributing samples more wisely: instead of sampling every offspring multiple times, \archivesampling{} re-samples only solutions already in the archive as they constitute the more promising solutions.
Thus, at every generation of \archivesampling{}, the archive is emptied, all solutions in it are reevaluated once and then added back to the archive using their moving-averaged fitness and descriptor. This new archive then undergoes a usual MAP-Elites loop. 

This periodical re-evaluation improves the knowledge on promising solutions, however, it might lead to drifting elites, i.e. elites moving to other cells. \archivesampling{} thus stores the $D$ best solutions per cell as a reservoir to avoid empty cells.
For a given cell, offspring are added until the depth $D$ is full; then, new offspring only replace lower-performing solutions.
The selection mechanism within these cells stays the same: only the best solution can be selected as a parent. 
The depth $D$ gives a second chance to unlucky solutions that have been evaluated with lower fitness than they could expect. Conversely, lucky solutions are rapidly re-evaluated and removed from the grid as necessary.  
Thus, \archivesampling{} addresses the main limitation of MAP-Elites in uncertain domains: its elitism. While MAP-Elites never question the content of the archive and favour lucky solutions; \archivesampling{} incrementally improves its certainty on solutions performance.

\begin{algorithm}[t!]
    \centering
    \caption{\eas{} (\ref{sec:eas})}\label{algo:eas}
    \begin{algorithmic}[1]
        \State \text{Initialise $\mathcal{A}$}
        \Repeat
             \vspace{2mm}
            \State \text{\textit{\# Evaluation number choice}}
            \State \text{$n_{evals}$ $\gets$ median\_evals\_number($\mathcal{A}$)}
            \vspace{2mm}
            \State \text{\textit{\# Archive reevaluation}}
            \State \text{$x \gets$ content of $\mathcal{A}$}
            \State \text{$\mathcal{A} \gets$ empty($\mathcal{A}$)}
            \State \text{$f$, $d$ $\gets$ reevaluations($x$, $N = 1$)}
            \State \text{$\mathcal{A} \gets$ add($\mathcal{A}$, $x$, $f$, $d$)}
            \vspace{2mm}
            \State \text{\textit{\# Offspring generation}}
            \State \text{$o \gets$ generate\_offspring($\mathcal{A}$)}
            \State \text{$f$, $d$ $\gets$ evaluations($o$, $N = n_{evals}$)}
            \State \text{$\mathcal{A} \gets$ add($\mathcal{A}$, $o$, $f$, $d$)}
            \vspace{2mm}
        \Until {reach convergence of $\mathcal{A}$}
    \end{algorithmic}
\end{algorithm}

\subsection{\eas{} (novel)} \label{sec:eas}


The \archivesampling{} approach still has a major limitation: solutions in the archive will gradually become more certain as they get reevaluated; however, any newcomer offspring with only one reevaluation could replace them. In short, \archivesampling{} compares solutions with different degrees of certainty. 
To overcome this, we propose a second algorithm: \eas{} (Alg.~\ref{algo:eas}). It relies on two mechanisms: (1) the archive-reevaluation mechanism (the same as \archivesampling{}) and (2) the offspring-reevaluation mechanism (introduced below).
The offspring-reevaluation mechanism (2) proposes to evaluate all new offspring $N$ times, with $N$ chosen based on the number of reevaluations of solutions already in the repertoire. 
Based on initial experiments, we take $N$ as the median number of evaluations of the solutions in the grid.
Due to (1), solutions in the archive get more and more reevaluated, so (2) leads to increase with time the number of samples spent on each offspring. 
Thus compared solutions have the same order of magnitude of “certainty”, adressing the main limitation of \archivesampling{}.

\eas{} can be seen as a version of Adaptive-Sampling for hardware-accelerated libraries. 
(1) reevaluates all elites, approximating the way Adaptive-sampling reevaluates contested elites during the offspring evaluation process. 
(2) can be seen as a proxy for the sampling mechanism of Adaptive Sampling, the main difference being that the number of samples is not offspring-dependent. 


\begin{table*}[ht]
\centering
  \caption{Task suite considered in this work.}
  \label{tab:tasks}
  \begin{tabular}{ c | c c c c }
  
    
    & \textsc{Arm} & \textsc{Hexapod} &
    \textsc{Walker} & \textsc{Ant} \\
    
    
    & \includegraphics[width = 0.18\textwidth]{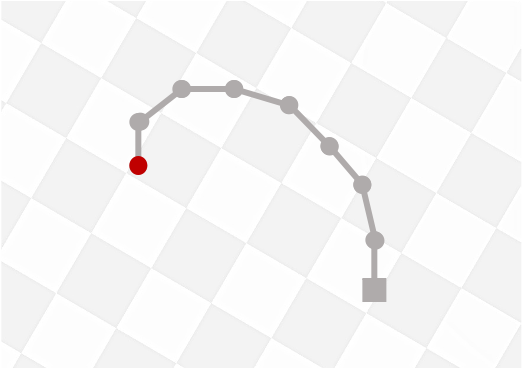} 
    & \includegraphics[width = 0.18\textwidth]{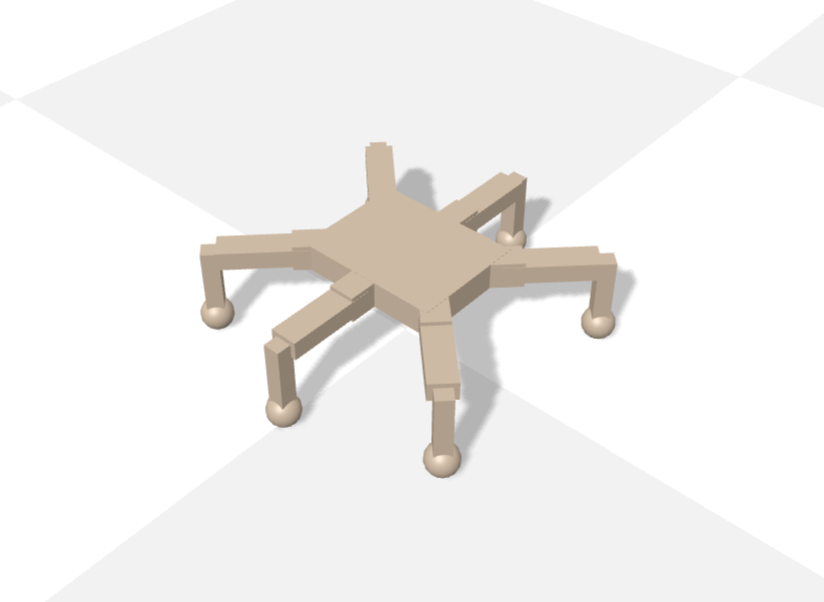} 
    & \includegraphics[width = 0.18\textwidth]{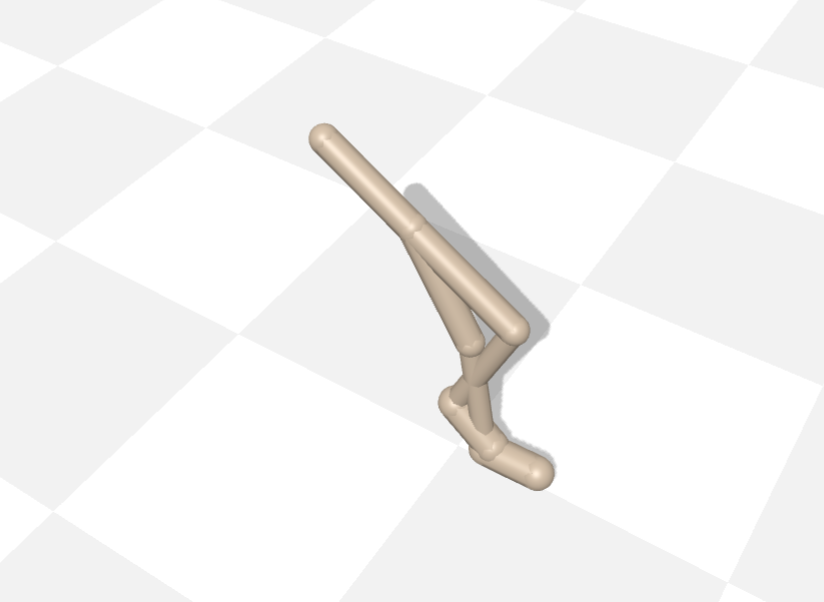} 
    & \includegraphics[width = 0.18\textwidth]{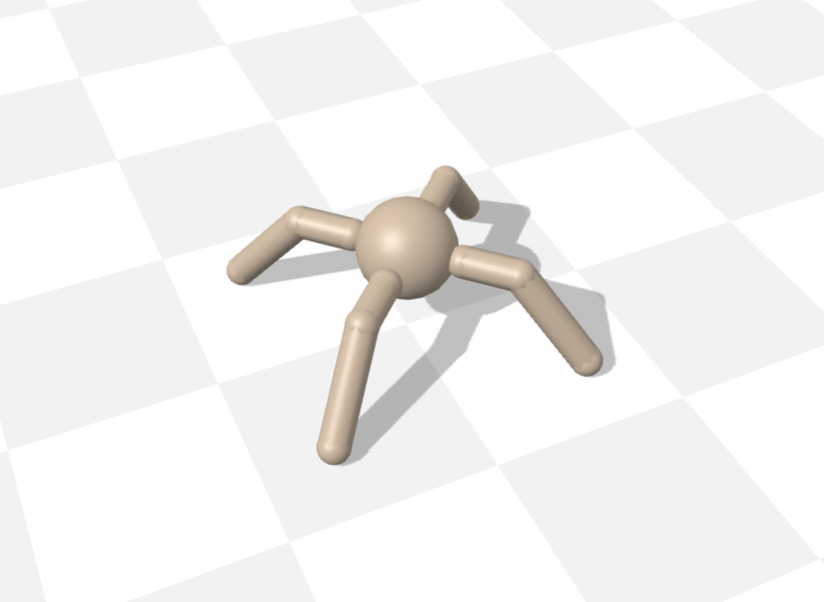} \\
    \addlinespace[0.05cm]
    
    \addlinespace[0.05cm]
    \midrule
    \addlinespace[0.05cm]
    
    \addlinespace[0.05cm]
    \textsc{Controller} 
    & \makecell{Desired joint positions.}
    & \makecell{Parameters of \\ periodic functions.}  
    & \makecell{Parameters of a \\ neural network.}
    & \makecell{Parameters of a \\ neural network.} \\

    
    \textsc{Dims} & $8$ & $36$ & $198$ & $296$ \\

    \textsc{Closed-loop} 
    & 
    & 
    & \checkmark
    & \checkmark \\

    \addlinespace[0.05cm]
    
    \addlinespace[0.05cm]
    \midrule
    \addlinespace[0.05cm]

    
    \addlinespace[0.05cm]
    \textsc{Fitness} 
    & \makecell{Negative variance \\ of joint positions.}
    & \makecell{Orientation error.}
    & \makecell{Accumulated forward progress, energy \\ usage penalty and surviving reward.}
    & \makecell{Energy usage penalty \\ and surviving reward.} \\
    \addlinespace[0.05cm]
    
    \addlinespace[0.05cm]
    \midrule
    \addlinespace[0.05cm]
    
    
    \addlinespace[0.05cm]
    \textsc{Descriptor} 
    & \makecell{End-effector \\ $[x, y]$ position.} 
    & \makecell{Final position of the robot's \\ center of mass $[x, y]$.}
    & \makecell{Proportion of time each foot of the \\ robot is in contact with the ground.}
    & \makecell{Final position of the robot's \\ center of mass $[x, y]$.} \\

    \textsc{Dims} & $2$ & $2$ & $2$ & $2$ \\

    \textsc{Nb Niches} & $1024$ & $1024$ & $1024$ & $1024$\\
    \addlinespace[0.05cm]
    
    \addlinespace[0.05cm]
    \midrule
    \addlinespace[0.05cm]
    
    
    \addlinespace[0.05cm]
    \textsc{Noise} 
    & \multicolumn{2}{c}{\makecell{Gaussian noise on fitness and descriptor.}}
    & \multicolumn{2}{c}{\makecell{Noise on initial joint positions and initial joint velocities.}} \\
    \addlinespace[0.05cm]
    
    
  \end{tabular}
  \label{fig:experiments}
\end{table*}

    
    
    
    
    
    
    

\section{\samplingsize{}: alternative for comparability} \label{sec:sampling_size}

Lim et al~\cite{qdax} studied the scalability of MAP-Elites to large batch-size (i.e. number of offspring generated per generation). 
In MAP-Elites, each solution is evaluated once so the batch-size is strictly equivalent to the number of evaluations per generation.
However, this is not the case for the methods considered in this work, which tend to perform multiple evaluations per solution. 
We study in this paper how the benefits offered by hardware-accelerated libraries can benefit QD in uncertain domains. 
As some algorithms perform multiple evaluations per offspring, we have to distinguish the number of evaluations performed in total by generation and the number of evaluations performed for each offspring. 
For this reason, we introduce the \textbf{\samplingsize{}}: the maximum allocated per-generation sampling budget.
For example, for a \samplingsize{} of $1024$, MAP-Elites would spend one evaluation on each offspring, thus generating $1024$ offspring solutions per generation. For the same \samplingsize{} of $1024$, a MAP-Elites-sampling approach spending $8$ samples on each offspring would only generate $128$ offspring per generation.
Using the \samplingsize{}, we can compare the performance of algorithms when given the same maximum number of samples per generation, making the comparison more meaningful.

\section{Experimental setup}

To facilitate later advances in \uqd{} we open-source our comparison code at \code{}.

\subsection{Tasks}

We perform our comparison on the tasks suite illustrated in Table ~\ref{fig:experiments}, taken from \cite{deepgrid, hexapod, benchmark} and chosen for their diversity of controller architectures, noise structures, and complexity.
Among these tasks, the Walker and Ant tasks are controlled in closed-loop and the Arm and the Hexapod in open-loop.
Closed-loop controllers use Deep Neural Networks and get information on the current states, so they have the ability to output correction to compensate for uncertainty. As a consequence, in the Walker and Ant tasks where the noise is applied on the initialisation, controllers are not all equally reproducible. These tasks present both the issue of performance estimation and reproducibility (see Sec.~\ref{sec:reproducibility}). 
On the contrary, open-loop controllers do not have access to the environment state, preventing them from compensating for uncertainty. Thus, the Arm and the Hexapod tasks are only subject to the performance estimation problem (see Sec.~\ref{sec:reproducibility}).

\begin{figure*}[t!]
\centering
\includegraphics[width = \hsize]{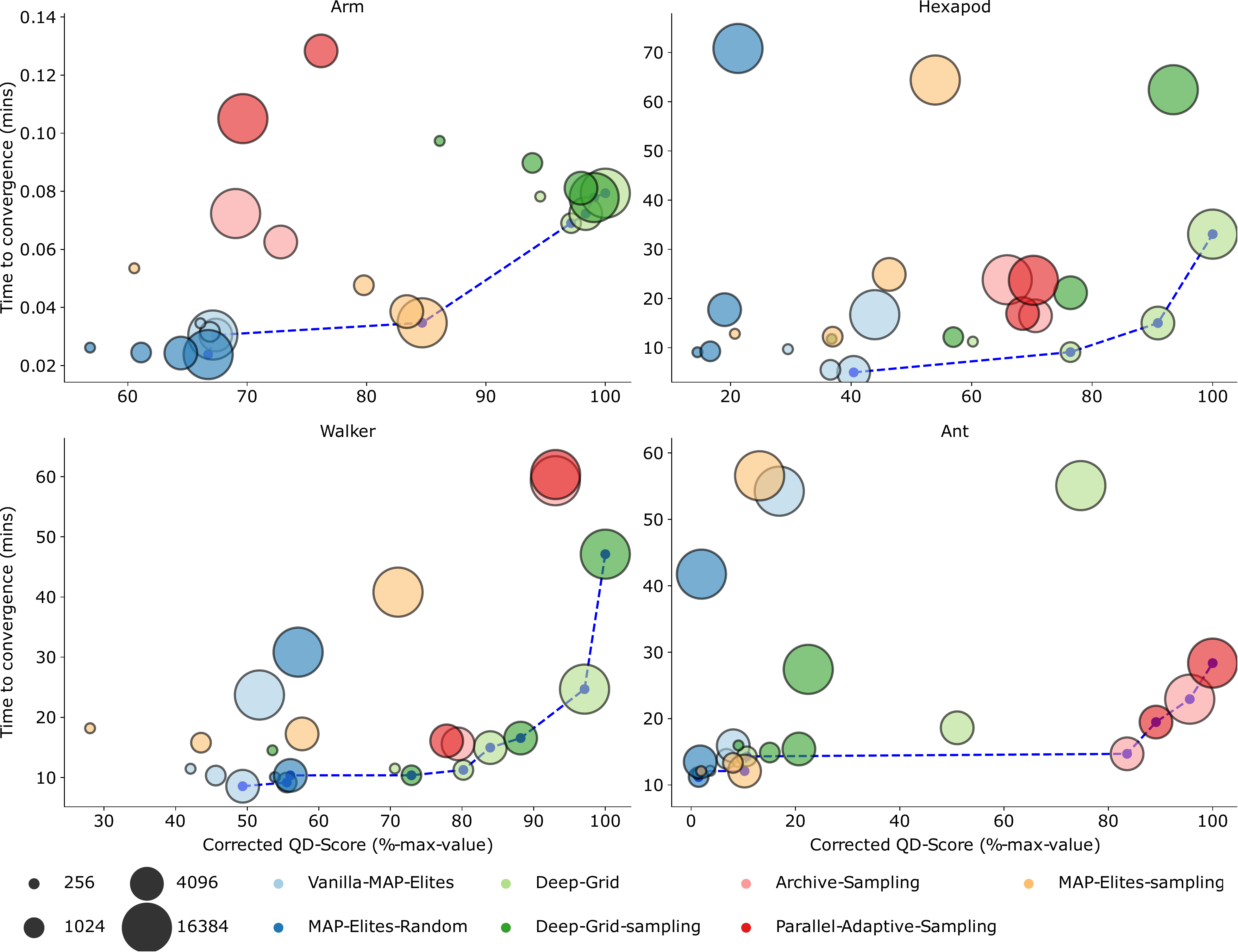}
\caption{
    Pareto front for all approaches on all tasks. 
    The x-axis is the Corrected QD-Score, quantifying the quality and diversity of the final collection; and the y-axis is the Time to convergence in minutes.
    For each approach, we represent increasing sampling-size as increasing marker-size. 
    The dashed blue line gives the Pareto-front.
    As detailed in Sec.~\ref{sec:baselines}, \archivesampling{} and \eas{} are not defined for values of sampling-size lower than the number of solutions in the archive.
    Each approach is replicated \replications{} times for each sampling-size value, each point corresponds to the median over replications.
}
\label{fig:pareto}
\end{figure*}

\subsection{Algorithms and baselines} \label{sec:baselines}

In the following, we compare the approaches from Sec.~\ref{sec:new_approaches} as well as \textbf{MAP-Elites} and \textbf{Random-MAP-Elites}. Random-MAP-Elites generates random solutions at each generation and add them to a MAP-Elites grid following the MAP-Elites addition condition.
Among the approaches from Sec.~\ref{sec:new_approaches}, \archivesampling{} and \eas{} reevaluate the entire content of the archive at each generation. For fairness, we count these evaluations as part of the  sampling-size: we deduct the maximum number of solutions in the archive from their sampling-size. Thus, these two approaches are undefined for values of sampling-size lower than the maximum number of solutions in the archive.
We replicate each algorithm \replications{} times on each task for each sampling-size, for a total of \totalreplications{} runs.
All approaches use CVT-shaped MAP-Elites archives \cite{cvt}.

\subsection{Metrics} \label{sec:metrics}

Our experimental analysis is divided into two parts: first, we compare the performance of the considered approaches; second, we study their ability to address the two problems raised in Sec.~\ref{sec:reproducibility}: Performance estimation and Reproducibility.

\subsubsection{Comparison in uncertain domains}
We quantify the performance of an algorithm using two metrics. 
First, the \textbf{Corrected QD-Score}, the QD-Score of the Corrected Archive described in Sec.~\ref{sec:uncertain_metrics}. 
Second, the \textbf{Time to convergence}, the wall-clock time, in seconds, needed to reach $95\%$ of the final Corrected QD-Score value. We estimate this time on the same hardware for all algorithms and replications. 

\subsubsection{Performance estimation and Reproducibility}
To quantify the ability of the algorithm to get good performance estimates, we use the \textbf{QD-Score loss}~\cite{pga_stochastic}: the difference between the QD-Score and the Corrected QD-Score, normalised by the QD-Score.
In addition, to assess the algorithm's ability to keep solutions with reproducible descriptors, we propose a new metric: the \textbf{Reproducibility-Score}. To compute it, we first get the descriptor variance of the $M$ reevaluations of each solution, normalised within each cell using the maximum observed-variance. This normalisation accounts for the difference in descriptor variance across the descriptor space.
The Reproducibility-Score is then computed as the sum over the archive of $ 1 - normalised\_variance$, to avoid penalising approaches that find more solutions.
One can similarly consider the fitness and define the Fitness-Reproducibility-Score.


\subsection{Hyperparameters} \label{sec:hyperparameters}

We run our experiments on Nvidia RTX6000 with $24GB$ of RAM, each run being allocated to one device only.
Most of the hyperparameters are common to all MAP-Elites variants, so we use the same values for all algorithms to guarantee comparability, provided in Appendix A. 
The algorithms from Sec.~\ref{sec:new_approaches} require two categories of hyperparameters that are not defined for MAP-Elites: number of samples $N$ and depth $D$. For each algorithm, we chose their optimal value independently, based on a study provided in Appendix C. 
In this study, we test $3$ possible values for each parameter, and choose the optimal values based on the comparison of $5$ replications of each algorithm on the full task suite across every sampling-size value, for a total of $\totalreplicationsparams$ runs.

\begin{figure*}[t!]
\centering
\includegraphics[width = \hsize]{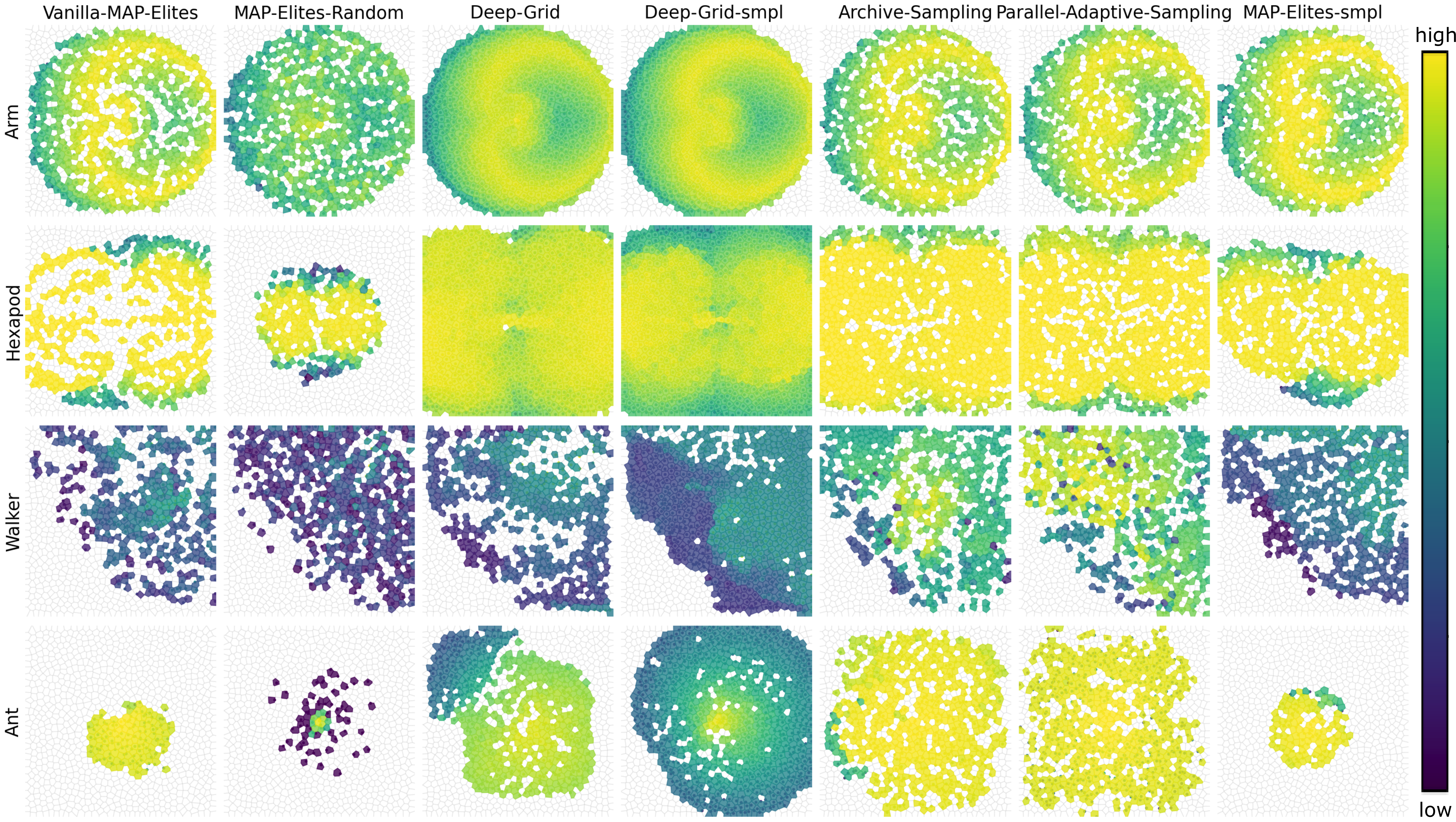}
\caption{
    Corrected Archives of all approaches on all tasks for sampling-size $16384$. We display one of the \replications{} replications, randomly chosen.
    Each cell represents one solution, with the colour representing the fitness (the brighter the better). The axis are the descriptor dimensions, defined in Table~\ref{tab:tasks}.
}
\label{fig:archives}
\end{figure*}

\section{Comparison in uncertain domains} \label{sec:results}

To display the trade-off between performance and convergence speed, we represent the metrics from Sec.~\ref{sec:metrics} as a Pareto plot:  the x-axis is the Corrected QD-Score, quantifying the quality and diversity of the final collection; and the y-axis is the Time to convergence. 
We also consider $4$ sampling-size values: $256, 1024, 4096, 16384$ and represent increasing sampling-size as increasing marker-size.
We display the plots in Fig.~\ref{fig:pareto}, with the dashed blue line giving the Pareto-front, and the corresponding archives in Fig.~\ref{fig:archives}. We report p-values based on the Wilcoxon rank-sum test with Bonferroni correction.

Lim et al.~\cite{qdax} prove that MAP-Elites' QD-Score scale with increasing batch-size. In the case of MAP-Elites, the sampling-size and batch-size are equivalent. Thus, we expect the Time to convergence (y-axis) to stay constant or decrease as the sampling-size (marker-size) increases. 
We observe that it is the case for sampling-size values: $256, 1024, 4096$. However, the Time to convergence increases again for $16384$ in all tasks requiring a simulator: Hexapod, Walker and Ant. This indicates that for $16384$, we reach the parallelisation plateau of our hardware and lose the time benefit of parallelisation. 
As we compare algorithms based on sampling-size, this limit is the same for all algorithms so does not impact our conclusions.

\subsection{Per-algorithm results}

As expected, the two baselines \textbf{MAP-Elites and MAP-Elites-Random} get lower QD-Score than \uqd{} approaches ($p < 5.10^{-5}$ for \eas{} and \archivesampling{} and $p < 5.10^{-2}$ for Deep-Grid). Still, at least one of them appears on the Pareto front in all tasks, thanks to their convergence speed. 
Interestingly, MAP-Elites-Random gets better Corrected-QD-Score than MAP-Elites in the Walker task ($p < 1.10^{-2}$). This illustrates the detrimental effect of uncertainty: when reevaluating the solutions kept by MAP-Elites, they prove to be sub-performing lucky solutions. 

\begin{figure*}[t!]
\centering
\includegraphics[width = \hsize]{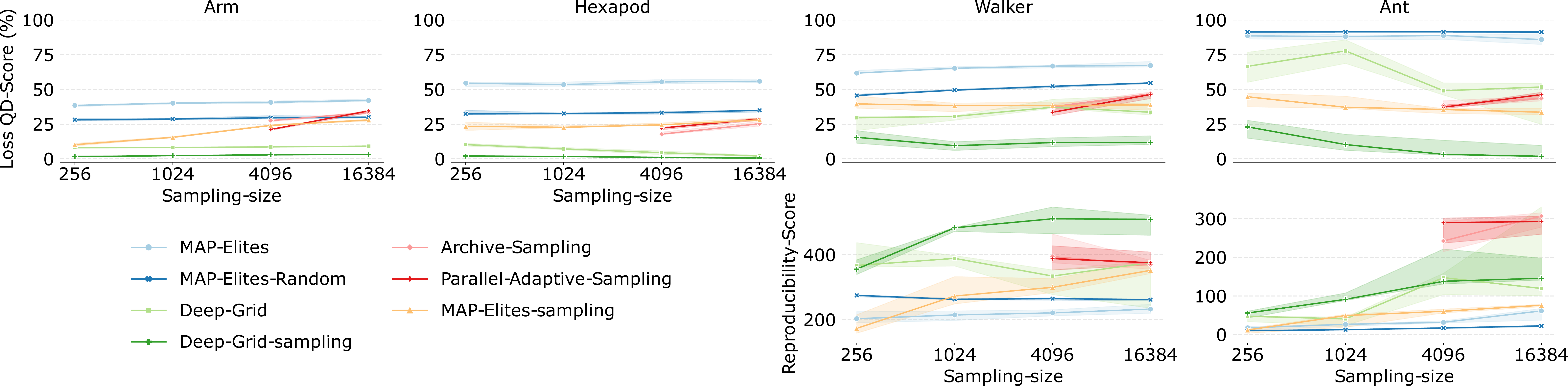}
\caption{ 
    Comparison with respect to sampling-size values of (top) QD-Score Loss, quantifying the estimations capability of the algorithm and (bottom) Reproducibility-Score, quantifying the reproducibility of solutions kept by the algorithm.
    All solutions in Arm and Hexapod are equally reproducible so we do not display the Reproducibility-Score for these tasks.
    As detailed in Sec.~\ref{sec:baselines}, \archivesampling{} and \eas{} are not defined for values of sampling-size lower than the number of solutions in the archive.
    Each algorithms is replicated \replications{} times for each sampling-size value.
}
\label{fig:loss}
\end{figure*}

\textbf{MAP-Elites-sampling} surprisingly does not perform that well. It is even dominated by MAP-Elites for sampling-size $256$ ($p < 5.10^{-2}$), and equivalent in Hexapod and Ant.
Its archives in these tasks (Fig.~\ref{fig:archives}) are also smaller, indicating that it discovers less diverse solutions than other approaches. 
This may be explained by an early convergence due to a lack of exploration. MAP-Elites-sampling retains only solutions that are reproducible from the start. However, early solutions are likely to have yet not learned how to walk in a reproducible manner but still constitute important stepping stones. 
By preventing these new brittle solutions from being considered as parents, MAP-Elites-sampling is preventing exploration. 
Previous work in EA applied to noisy domains already highlighted that static explicit-sampling might induce a lack of exploration \cite{ea_uncertain_2}.
Thus, in \uqd{}, it seems more promising to add diverse solutions to the archive and later question them, rather than only retaining highly reproducible solutions from the start.

\textbf{Deep-Grid} is on the Pareto-front in three tasks out of four for almost all sampling-sizes, making it the best performing approach. These results highlight the effectiveness in the considered tasks of using past solutions to approximate distributions.
Deep-Grid only struggles in the Ant task, which appears to be the most challenging of the suite, as it is $3D$ and closed-loop controlled while having the largest search-space. 
Flageat et al.~\cite{pga_stochastic} already highlighted that Deep-Grid mechanisms struggle more than sampling-based ones when applied to large-dimension search-spaces. 

\textbf{Deep-Grid-sampling} is equivalent to Deep-Grid for large sampling-size values in Arm and Walker but get lower Corrected QD-Score for small ones ($p < 1.10^{-3}$).
As could be expected from MAP-Elites-sampling's results, Deep-Grid-sampling also worsens Deep-Grid Corrected QD-Score in Hexapod ($p < 5.10^{-4}$) and Ant ($p < 5.10^{-2}$) where it likely prevents exploration, as can also be observed in the archives in Fig.~\ref{fig:archives}. Interestingly, Deep-Grid-sampling converges to its final sub-optimal archive faster than every other approaches in Ant with $16384$ ($p < 5.10^{-4}$), which corroborates the hypothesis of stucked exploration.
Overall, Deep-Grid relies on filling cells with reproductions of similar solutions. Introducing sampling reduces the total number of solutions added to the cells at each generation, slowing down these mechanisms. Thus, Deep-Grid-sampling does not improve performance in tasks where Deep-Grid prove efficient, such as Arm or Hexapod. 

\textbf{\archivesampling{} and \eas{}} get good results in Hexapod and Walker although they are dominated by Deep-Grid. 
However, they outperform every other approach in Ant ($p < 5.10^{-4}$ for Corrected QD-Score), which, as highlighted earlier, constitutes the most challenging task.
\archivesampling{} and \eas{} take the best of both worlds: they keep early, promising solutions, but also question solutions over time to ensure their performance.
Thus, they leverage the available per-generation sampling budget by distributing samples wisely while allowing exploration and constantly questioning solutions, which prove promising in the Ant task.
The additional offspring-reevaluation mechanism in \eas{} allows it to get slightly higher Corrected QD-Score values than \archivesampling{} but to a higher time cost. Thus, neither of the two approaches dominates the other.

\subsection{Conclusion}

Overall, from these results, Deep-Grid appears the optimal \uqd{} approach in Arm, Hexapod and Walker whatever the sampling-size. It also proves optimal for small sampling-size in Ant. 
However, as soon as the sampling-size becomes large enough to afford \archivesampling{} or \eas{} in Ant, they become by far the most interesting approaches. 
Ant is the most challenging task of our benchmark: it is closed-loop unlike the Hexapod and $3$-dimensional, unlike the Walker. 
Thus, the good performance of \archivesampling{} and \eas{} on this task compared to all other approaches makes them promising, especially as they also get good results in Hexapod and Walker.

This study also demonstrates once again the limitations of MAP-Elites in \uqd{} domains. 
It also highlights that the simple MAP-Elites-sampling algorithm is not necessarily a good approach in uncertain domains, as it seems to prevent exploration, corroborating previous results in EA \cite{ea_uncertain_2}. 

\section{Analysis of performance estimation and Reproducibility} \label{sec:ablation}

We display the metrics from Sec.~\ref{sec:metrics} in Fig.~\ref{fig:loss} and Fig.~\ref{fig:archives_var}. We also give the Fitness-Reproducibility-Score results in Appendix D. 
All solutions in the Arm and Hexapod tasks are equally reproducible as the uncertainty is a Gaussian noise on the fitness and descriptor that is the same for all solutions. Thus, we do not display the Reproducibility-Score for these two tasks.
We report p-values based on the Wilcoxon rank-sum test with Bonferroni correction.

\begin{figure*}[t!]
\centering
\includegraphics[width = \hsize]{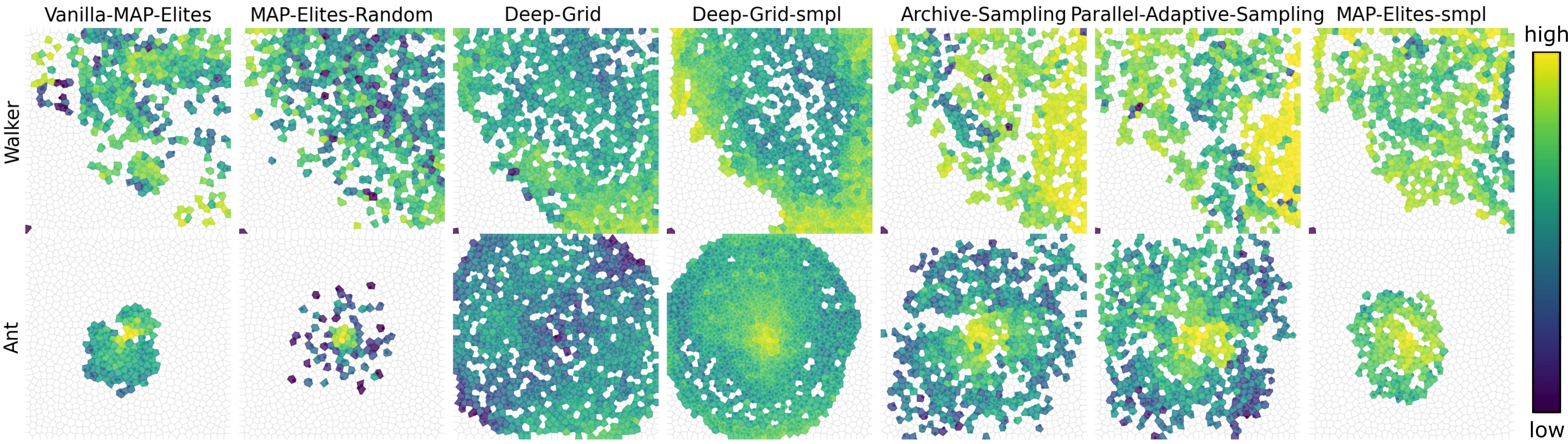}
\caption{
    Per-cell Reproducibility-Score of all approaches for sampling-size $16384$.
    We display one of the \replications{} replications, randomly chosen.
    We give the archives in Walker and Ant, but not in Arm and Hexapod as all solutions have the same variance.
    Each cell represents one solution, with the colour representing its Reproducibility-Score, which corresponds to $1 - normalised\_variance$ (the brighter the higher). The axis are the descriptor dimensions, defined in Table~\ref{tab:tasks}.
}
\label{fig:archives_var}
\end{figure*}

\subsection{Performance-estimation capability (QD-Score loss)}

The QD-Score loss quantifies the inaccuracy of the estimations of fitness and descriptor done by the algorithms. As it is a loss, algorithms should seek to minimise it. 

According to this metric, \textbf{MAP-Elites} is consistently the worst of the considered approaches. 
Even \textbf{MAP-Elites-Random} is significantly better ($p < 5.10^{-4}$), except on the Ant task.
MAP-Elites relies on improving existing solutions, which leads to evaluating repetitively similar solutions until they get lucky evaluations. This is not the case with MAP-Elites-Random, which always considers new random solutions and thus, on average, better estimates their performance.
All explicit-sampling approaches: \textbf{MAP-Elites-sampling}, \textbf{\archivesampling{}} and \textbf{\eas{}} get equivalent losses values, which is consistent with the fact that they all rely on using the averages over samples as estimates.
However, the implicit-sampling approaches (i.e. Deep-Grid approaches) get better approximation than explicit-sampling ones on open-loop tasks ($p < 1.10^{-3}$), while their approximations are equivalent in Walker and Ant.
\textbf{Deep-Grid} approximates the distribution within a cell using multiple solutions, while explicit-sampling approaches use the average of their samples. 
It seems that such approximation works well for simple Gaussian noise distribution as considered in the Arm and Hexapod but quickly becomes intractable for more complex distributions as in the Walker and Ant. 
Interestingly, \textbf{Deep-Grid-sampling}, which combines these two mechanisms, manages to get the lowest loss of all approaches ($p < 5.10^{-3}$ except for MAP-Elites-sampling in Ant where $p < 2.10^{-2}$).


\subsection{Reproducibility (Reproducibility-Score)}

We propose the Reproducibility-Score to quantify the reproducibility of solutions selected by each algorithm. 
Algorithms should seek to maximise it. 

According to this metric, \textbf{MAP-Elites-Random}, and \textbf{MAP-Elites} find less reproducible solutions than every other approach.
Engebraaten et al.~\cite{glette_stochastic} highlighted that MAP-Elites has a bias favouring solutions with large Descriptor-variance, as they tend to easily populate multiple cells and thus to better survive in the population.
MAP-Elites-Random optimises solutions less than MAP-Elites and thus suffers less from this bias in the Walker task ($p < 5.10^{-2}$), corroborating the results from the QD-Score Loss.
However, in Ant, solutions with large variance have a higher chance of populating hard-to-reach cells, and would hardly be questioned as MAP-Elites-Random does not optimise solutions any further. Thus, MAP-Elites-Random selects less reproducible solutions than MAP-Elites in Ant ($p < 5.10^{-4}$).

Sampling is an efficient way to counteract this bias, as it reduces the area that these large-descriptor-variance solutions can achieve. 
This is what we observed from the results of \textbf{\archivesampling{}} and \textbf{\eas{}}, which find significantly more reproducible solutions than MAP-Elites and MAP-Elites-Random on both tasks ($p < 1.10^{-3}$).
However, MAP-Elites-sampling, which also relies on sampling, find solutions that are equivalently reproducible to those of MAP-Elites and MAP-Elites-Random. Again, this seems to be due to its lack of exploration: MAP-Elites-sampling likely finds more-reproducible solutions, but fewer of them, giving an equivalent Reproducibility-Score.

Alternatively, the questioning mechanism of \textbf{Deep-Grid} also proves promising as it finds more reproducible solutions than MAP-Elites and MAP-Elites-Random  ($p < 2.10^{-2}$ except for Walker with high-sampling size).
This mechanism replaces every solution in a cell with equal probability, leading to slowly getting rid of solutions with large variance, as they do not end up repetitively in the same cell.
As observed with the QD-Score loss, \textbf{Deep-Grid-sampling} combine these two ideas and manage to find the more reproducible solutions in the Walker task ($p < 5.10^{-2}$ for sampling size 256 and all approaches except Deep-Grid and $p < 1.10^{-3}$ for all other sampling-size). 
However, it proves slightly less good in the more complex Ant, where it gets outperformed by \archivesampling{} and \eas{} ($p < 1.10^{-2}$).


\subsection{Conclusion}

This section aims to understand better the underlying dynamics of the \uqd{} approaches studied in this work. 
We show that Deep-grid approximates the expected fitnesses and descriptors of solutions well in domains with simple noise structure. However, its distribution approximation seems to struggle with more complex noise structures. In addition, Deep-Grid tends to find less reproducible solutions than Deep-Grid-sampling, \archivesampling{} and \eas{}.
On the contrary, these three approaches succeed in finding highly reproducible solutions while getting consistently reliable fitness and descriptor estimates across tasks. 
However, previous sections prove that Deep-Grid-sampling is underperforming in term of QD-Score in the complex Ant and Walker tasks. 
Thus, this last point combined with their good QD-Score makes \archivesampling{} and \eas{} strong competitors in complex closed-loop \uqd{} domains.

\section{Conclusion and Discussion}

In this work, we propose \uqd{}, a framework that aims to formalise the problem of QD in uncertain domains, encountered multiple times across QD literature \cite{deepgrid, glette_stochastic, adaptive, pga_stochastic}. 
\uqd{} is a special case of QD optimisation in which fitness and descriptors for each solution are no longer fixed values but rather a distribution over possible values.
Building on this first contribution, we also propose a new methodology to evaluate \uqd{} approaches, composed of a set of tasks and metrics, as well as a metrics-computation procedure.
As third and fourth contributions, we propose three new approaches: \archivesampling{}, \eas{} and Deep-grid-sampling and compare them to existing \uqd{} approaches. 
We empirically show that the best performing \uqd{} approach is Deep-Grid for simple uncertain domains, as it proves fast and high performing.
However, we also demonstrate the benefit of \archivesampling{} and \eas{} for more complex applications.
These approaches slightly underperform Deep-Grid on simple domains but exhibit competitive results on complex tasks. More importantly, they have the interesting property to select reproducible solutions, making them very promising for more realistic tasks.
Our study displays once again the limitations of MAP-Elites in \uqd{} tasks, but also demonstrates that simple sampling-based approaches are not that interesting in such settings as they lead to a lack of exploration.
We hope our framework and method will constitute a meaningful benchmark for later works considering \uqd{}.

\section*{Acknowledgments}

This work was supported by the Engineering and Physical Sciences Research Council (EPSRC) grant EP/V006673/1 project REcoVER. We also want to thank the members of the Adaptive and Intelligent Robotics Lab at Imperial College London for their useful comments and in particular Jenny Zhang for her work on the Deep-Grid algorithm.

\bibliographystyle{IEEEtran}
\bibliography{bibliography}

\section*{Biography Section}

\begin{IEEEbiography}[{\includegraphics[width=1in,height=1.25in,clip,keepaspectratio]{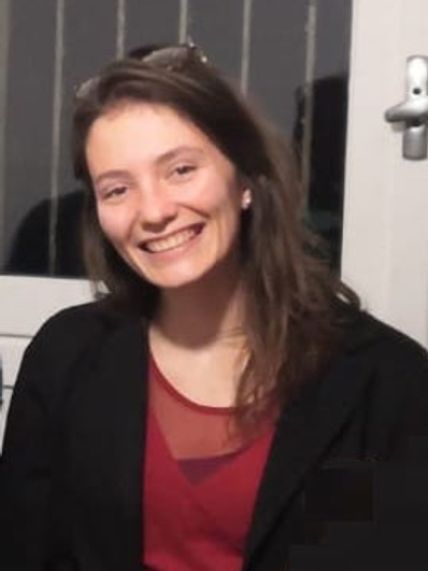}}]{Manon Flageat}
Manon Flageat received the M.Sc. degree in Biomedical Engineering (specialism: Neurotechnology) from Imperial College London (United Kingdom) in 2019, and the engineering degree from the École Nationale Supérieure des Mines de Saint-Étienne, France, in 2019. 
She is currently pursuing the Ph.D. degree in learning algorithms for robotics with the Adaptive and Intelligent Robotics Laboratory, Department of Computing, Imperial College London.
Her research focuses on Quality-Diversity algorithms, in particular applied to uncertain environment, as well as Deep Reinforcement Learning and synergies between these two types of learning algorithms.
She is currently also a Teaching Scholar in the Department of Computing, Imperial College London. 
\end{IEEEbiography}

\begin{IEEEbiography}[{\includegraphics[width=1in,height=1.25in,clip,keepaspectratio]{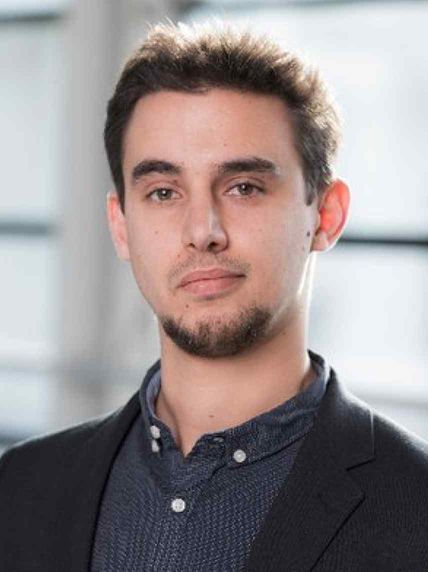}}]{Antoine Cully}
Antoine Cully is Senior Lecturer (Associate Professor) at Imperial College London (United Kingdom) and the director of the Adaptive and Intelligent Robotics Lab. His research is at the intersection between artificial intelligence and robotics. He applies machine learning approaches, like evolutionary algorithms, on robots to increase their versatility and their adaptation capabilities. In particular, he has recently developed Quality-Diversity optimization algorithms to enable robots to autonomously learn large behavioural repertoires. For instance, this approach enabled legged robots to autonomously learn how to walk in every direction or to adapt to damage situations.
Antoine Cully received the M.Sc. and the Ph.D. degrees in robotics and artificial intelligence from the Sorbonne Université in Paris, France, in 2012 and 2015, respectively, and the engineering degree from the School of Engineering Polytech’Sorbonne, in 2012. His Ph.D. dissertation has received three Best-Thesis awards. He has published several journal papers in prestigious journals including Nature, IEEE Transaction in Evolutionary Computation, and the International Journal of Robotics Research. His work was featured on the cover of Nature (Cully et al., 2015), received the "Outstanding Paper of 2015" award from the Society for Artificial Life (2016), the French "La Recherche" award (2016), and two Best-Paper awards from GECCO (2021, 2022).
\end{IEEEbiography}

{\appendices

\section{Hyperparameters} \label{app:hyperparam}

All the algorithms considered in this work are based on Vanilla-MAP-Elites with line mutation from \cite{line} (except for MAP-Elites-Random which does not require any mutation). We give in Tab.~\ref{tab:hyperparams} the values of the hyperparameters used for all algorithms.
As our comparison is based on sampling-size (i.e. generation-based sampling-budget), each algorithm generates the maximum number of offspring per generation so as to not overcome this budget. Thus, they all generate a different number of offspring per generation.

\begin{table}[ht]
  \centering
  \caption{Hyper-parameters used for all approaches in this work.}
  \label{tab:hyperparams}
  \begin{tabular}{c c}
    
    \textsc{Parameter} & \textsc{Value} \\
    
    \addlinespace[0.05cm]
    \midrule
    \addlinespace[0.05cm]
    
    Number of descriptor niches & $1024$ \\
    Line mutation param. $\sigma_1$ & $0.005$ \\
    Line mutation param. $\sigma_2$ & $0.05$ 

\end{tabular}
\end{table}

We give in Tab.~\ref{tab:hyperparams_tasks} additional task hyperparameters. 
We report in the paper the real-time convergence, in minutes. To compute these times, we run each algorithm for a maximum number of generations chosen sufficiently high for each task and compute the time needed to reach $95\%$ of the final Corrected QD-Score value.
In the paper, we also report algorithms' archives, we give in Tab.~\ref{tab:hyperparams_tasks} the extreme fitness and descriptor-variance values used for these plots.

\begin{table}[ht]
\centering
  \caption{Hyperparameters for each task.}
  \label{tab:hyperparam_tasks}
  \begin{tabular}{ c c c c c }

    &
    \textsc{Arm} & 
    \textsc{Hexapod} &
    \textsc{Walker} & 
    \textsc{Ant} \\
    
    \addlinespace[0.05cm]
    \midrule
    \addlinespace[0.05cm]
    
    Max generations & 2000 & 3000 & 2000 & 3000 \\
    Simulation length (s) & - & 5 & 8 & 5 \\
    Min Fitness & -0.24 & -745 & -85 & -750 \\
    Max Fitness & 0.00027 & -0.079 & 2328 & 250 \\
    Min Descriptor-var & - & - & 0.00043 & 0.025 \\
    Max Descriptor-var & - & - & 0.31 & 14 \\


  \end{tabular}
  \label{tab:hyperparams_tasks}
\end{table}

In Tab.~\ref{tab:hyperparams_noise}, we give the structure and amplitude of noise used in the different tasks. 
The Walker and Ant tasks are standard benchmarks in Reinforcement Learning \cite{benchmark}, we used the noise already implemented in the Brax environments \cite{brax} and commonly used in previous works in this field. 

\begin{table}[ht]
\centering
  \caption{Noise distribution used in each task.}
  \label{tab:hyperparam_noise}
  \begin{tabular}{ c c c c }
    
    \textsc{Arm} & 
    \textsc{Hexapod} &
    \textsc{Walker} & 
    \textsc{Ant} \\
    
    \addlinespace[0.05cm]
    \midrule
    \addlinespace[0.05cm]
    
    \multicolumn{2}{c}{\makecell{Noise on fitness and descriptor.}}
    & \multicolumn{2}{c}{\makecell{Noise on initial joint positions $p$ \\ and initial joint velocities $v$.}} \\
    \addlinespace[0.05cm]
    
    \makecell{$f \sim \mathcal{N}(0.01)$ \\ $d \sim \mathcal{N}(0.01)$}
    & \makecell{$f \sim \mathcal{N}(0.05)$ \\ $d \sim \mathcal{N}(0.05)$} 
    & \makecell{$p \sim \mathcal{U}(0.005)$ \\ $v \sim \mathcal{U}(0.005)$}
    & \makecell{$p \sim \mathcal{U}(0.1)$ \\ $v \sim \mathcal{N}(0.1)$} \\
    
  \end{tabular}
  \label{tab:hyperparams_noise}
\end{table}

\section{Performance estimation for metrics computation} \label{app:reevalchoice}

\begin{figure}[h]
\centering
\includegraphics[width = \hsize]{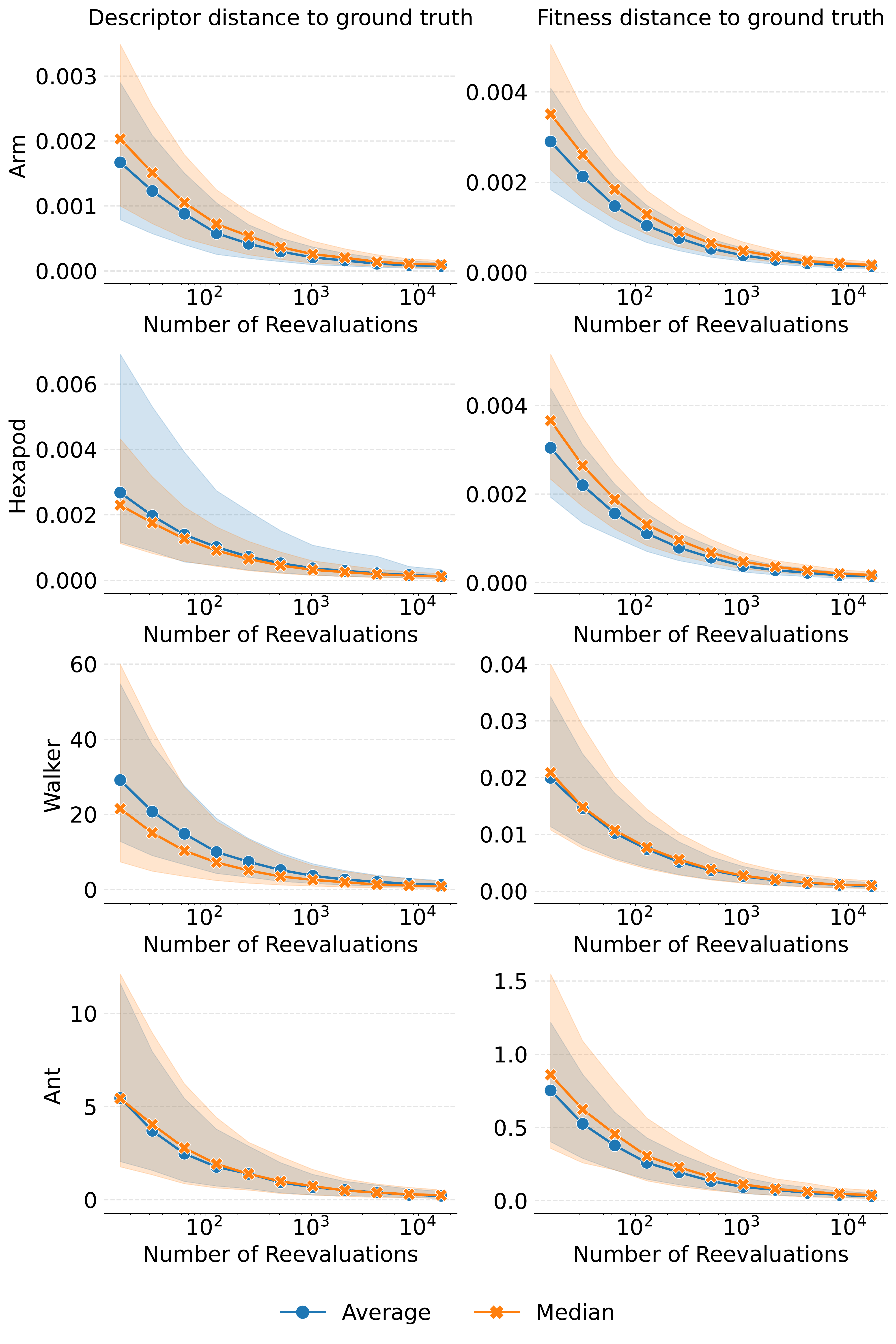}
\caption{
    Study of the number of reevaluations and metrics for reevaluations.
    The x-axis is the number of reevaluations $M$ and the y-axis is the distance of the average (blue) or median (orange) to the ground-truth average or median respectively; in descriptor-space (left) or in fitness-space (right).
    The plain line gives the median over final solutions and \replications{} replications of MAP-Elites. The shaded area corresponds to the quartiles.
}
\label{fig:reeval-tunning}
\end{figure}

As detailed in Sec.~III.D, previous works in \uqd{} have computed metrics estimations based on an arbitrarily-chosen number of reevaluations $M = 50$ for each solution in the archive.
In addition, they have also based their estimations on averaging these $M$ re-evaluations. However, Doerr and Sutton~\cite{median} suggested that the median is a more reliable estimator for metric estimation in uncertain domains.
Thus, in this section, we propose a study of the value of $M$ as well as a comparison using median and average as estimates. We aim to provide meaningful rationales for these two choices. 

One main limitation in uncertain settings, in particular closed-loop uncertain settings, is that the ground-truth value of the fitness and descriptor is not easily accessible, not even to an expert. It is thus difficult to provide an oracle to compare the estimation to. 
We propose to use a number of reevaluations $M_{max}$ sufficiently high to be considered a good estimation of the ground-truth. 
We then aim to find the smaller possible value of $M$ allowing us to get a good estimate of this ground-truth value. 
We apply the following process:
\begin{enumerate}
    \item For each domain, we run \replicationsreeval{} replications of MAP-Elites.
    \item For each replication, we perform $M_{max}$ replications of each solution in the final archive. We then compute $f^{gt}_{avg}$ and $d^{gt}_{avg}$ as the average of the $M_{max}$ replications, where $gt$ stands for "ground-truth". We do the same for $f^{gt}_{med}$ and $d^{gt}_{med}$, median over $M_{max}$ replications.
    \item For each candidate value of $M$: 
    \begin{enumerate}
        \item For each solution, we perform $M$ replications and compute $f_{avg}$, $d_{avg}$, $f_{med}$ and $d_{med}$ as the average and median over the $M$ replications.
        \item For each solution, we compute the distance between $f_{avg}$ and $f^{gt}_{avg}$, $d_{avg}$ and $d^{gt}_{avg}$, $f_{med}$ and $f^{gt}_{med}$, $d_{med}$ and $d^{gt}_{med}$ as the estimation error for $M$ replications.
    \end{enumerate} 
\end{enumerate}
We propose to use $M_{max} = 16384$ and candidate values for $M$: $16, 32, 64, 128, 256, 512, 1024, 2048, 4096, 8192, 16384$. 
We give the results of the study in Fig.~\ref{fig:reeval-tunning}.
Our first observation is on the choice of $M_{max}$. In our study, we are performing the same steps for $M = M_{max}$ as for other values of $M$. The data-point for $M = M_{max} = 16384$ in Fig.~\ref{fig:reeval-tunning} shows that the error between two distinct $16384$ re-evaluations is close to $0$. This is a good indicator that this value is high enough to be considered ground-truth. 
The results in Fig.~\ref{fig:reeval-tunning} also indicate that the error has converged for values of $M$ greater than $512$. As we are looking for the smallest possible value of $M$ to get a good estimate of the ground-truth, we pick $M=512$.
Our study does not highlight major differences between using median and average over the $M$ replications. However, we chose to stick with the median for consistency with \cite{median}.

In this study, to compute the average (or median) descriptor, we compute the average (or median) over each dimension of the descriptor and concatenate them. This approach is the one used in previous QD works for uncertain domains \cite{adaptive, deepgrid, pga_stochastic, glette_stochastic}.
However, this gives a solution that might be really different from the existing solutions if the considered population is multimodal.
An alternative would be to use the geometric median over descriptor dimensions \cite{geometric_median}, but it is approximated using iterative methods, and using it in our context would slow down a lot the reevaluation process. 
We leave this improvement for future work.

\section{Reproducibility-Scores} \label{app:fitness_var}

To complement the results from Sec.~IX, we display in Fig.~\ref{fig:fitness_var} the Fitness-Reproducibility-Score, and in Fig.~\ref{fig:fitness_var_archive} the corresponding archives. 
Both the Fitness and Descriptor variance are normalised per cell, we also provide in Fig.~\ref{fig:max_var_archive} the maximum observed-variances used for this normalisation.

\begin{figure}[h]
\centering
\includegraphics[width = \hsize]{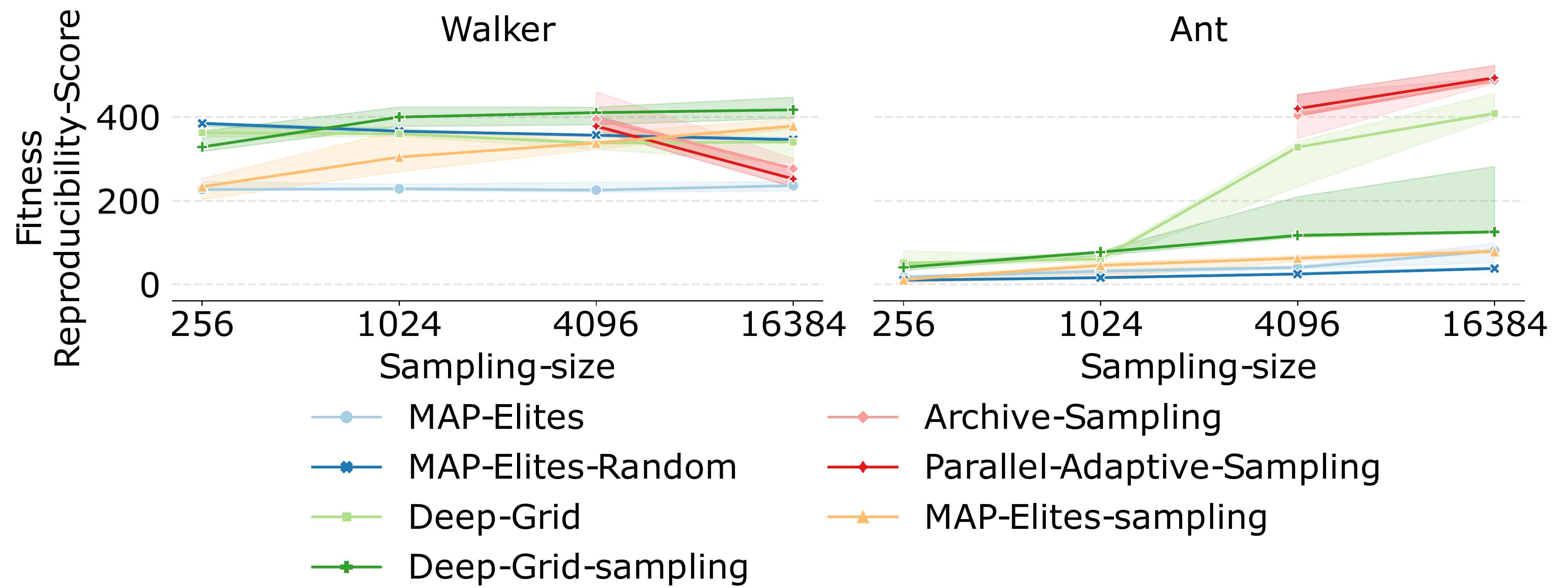}
\caption{
    Comparison of all approaches on the Walker and Ant tasks with respect to sampling-size values, according to Fitness-Reproducibility-Score, quantifying the reproducibility of solutions kept by the algorithm.
    Each algorithms is replicated \replications{} times for each sampling-size value.
}
\label{fig:fitness_var}
\end{figure}

\begin{figure}[t!]
\centering
\includegraphics[width = 0.65\hsize]{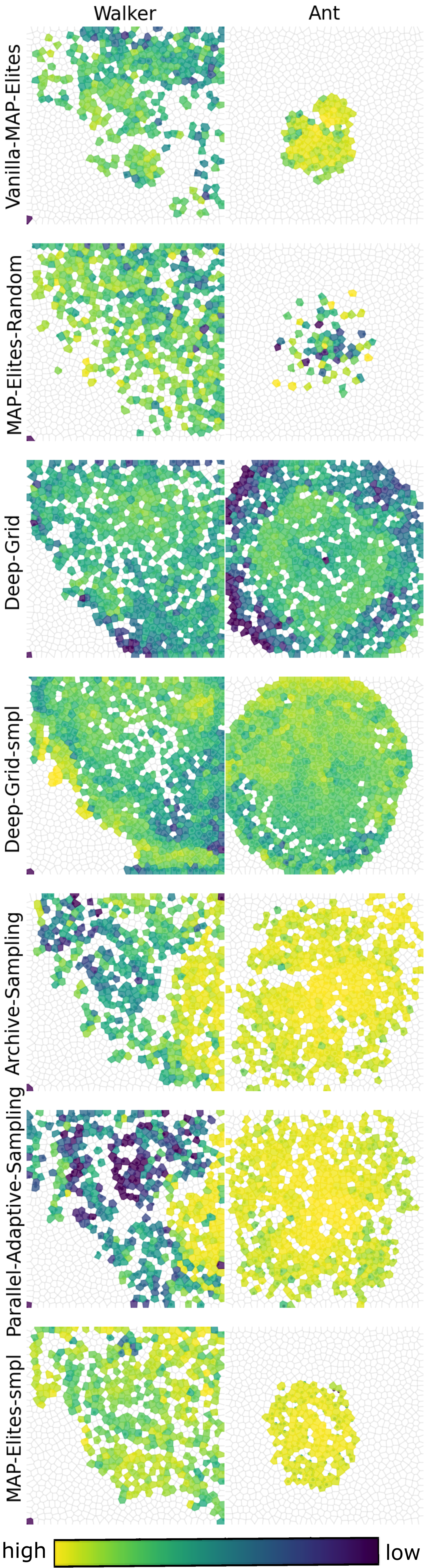}
\caption{
    Per-Cell Fitness-Reproducibility-Score of all approaches for sampling-size $16384$.
    We display one of the \replications{} replications, randomly chosen.
    We give the archives in Walker and Ant, but not in Arm and Hexapod as all solutions have the same variance.
    Each cell represents one solution, with the colour representing its Reproducibility-Score, which corresponds to $1 - normalised\_variance$ (the brighter the higher). The axis are the descriptor dimensions, defined in Table II.
}
\label{fig:fitness_var_archive}
\end{figure}

\begin{figure}[t!]
\centering
\includegraphics[width = \hsize]{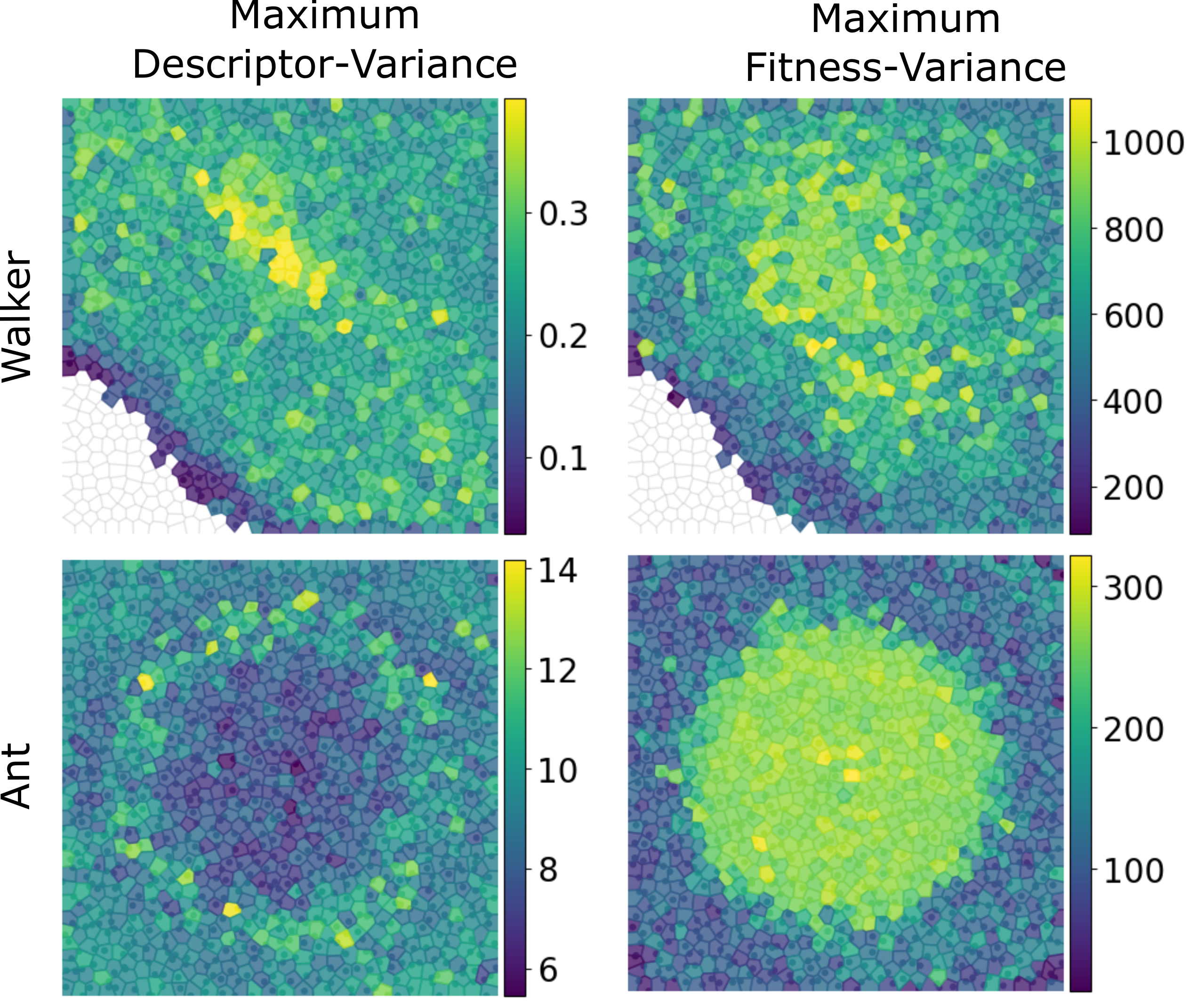}
\caption{
    Maximum observed Descriptor-variance (left) and maximum observed Fitness-variance (right) across all \replications{} replications of each algorithm for all sampling-size.
    We give the archives in Walker and Ant, but not in Arm and Hexapod as all solutions have the same variance.
    Each cell represents one solution, with the colour representing its variance (the brighter the higher, so the darker the better). The axis are the descriptor dimensions.
}
\label{fig:max_var_archive}
\end{figure}

\section{Notes on metric computation in \uqd{}} \label{app:deepgrid}

As detailed in the main paper, when computing the Corrected metrics, the $M$ reevaluations of each cell are done by sampling each cell $M$ times using its in-cell selector. 
Among the approaches considered in this paper, only Deep-Grid and Deep-Grid-sampling have such in-cell selector that differ from MAP-Elites.
This in-cell selector is an important component of the Deep-Grid algorithm, as Deep-Grid returns a deep archive, which depth aims to be taken into account for later usage. 
For fairness, we provide in Fig.~\ref{fig:best_pareto} an alternative comparison, where we use the best solution of each cell to compute metrics for all approaches.
While Deep-Grid and Deep-Grid-sampling get slightly lower Corrected QD-Score in this comparison, the results lead to similar conclusion than the main results.

\begin{figure}[h]
\centering
\includegraphics[width = 0.88\hsize]{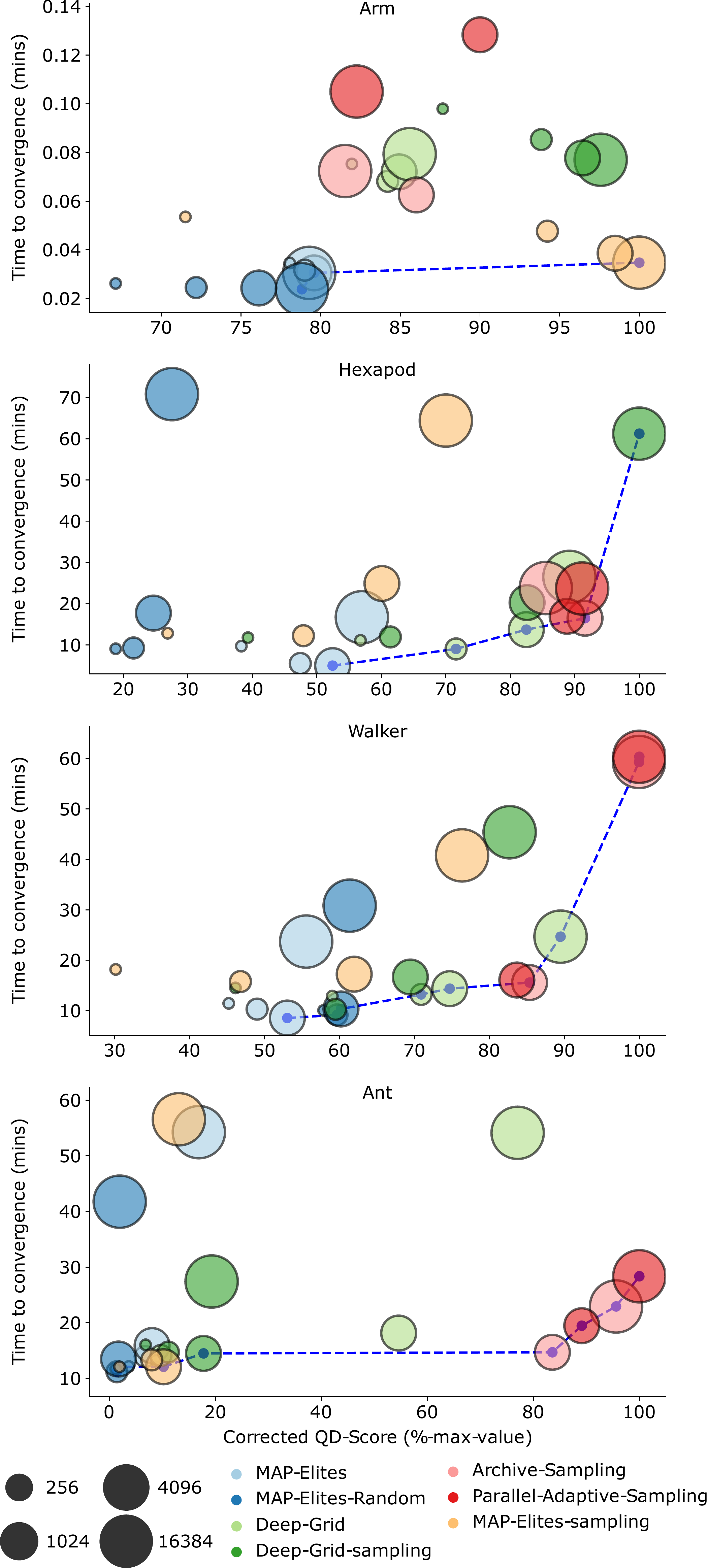}
\caption{
    Pareto front for all approaches on all tasks using only the best final solution for all approaches. 
    The x-axis is the Corrected QD-Score, quantifying the quality and diversity of the final collection; and the y-axis is the Time to convergence in minutes.
    For each approach, we represent increasing sampling-size as increasing marker-size. 
    The dashed blue line gives the Pareto-front.
    As detailed in Sec.~\ref{sec:baselines}, \archivesampling{} and \eas{} are not defined for values of sampling-size lower than the number of solutions in the archive.
    Each approach is replicated \replications{} times for each sampling-size value, each point corresponds to the median over replications.
}
\label{fig:best_pareto}
\end{figure}

\section{Notes on \archivesampling{} and \eas{} implementation} \label{app:archivesampling}

As detailed in the main paper, \archivesampling{} and \eas{} have two main phases: (1) the archive-reevaluation and (2) the offspring-evaluation. The main difference between these two algorithms is that \eas{} change the number of evaluations $N$ performed in steps (2). 
In our implementation, to take advantage of the hardware-accelerated context, these two steps are performed in parallel and not sequentially: the content of the archive is concatenated with $N$ copies of the offspring and everything is evaluated once, before being added to an empty archive.
At the level of the algorithm, this is equivalent to performing the archive-reevaluation first and the offspring-evaluation, except for the first generation. To account for this, the archive-reevaluation starts at the second generation.

\section{Parameters of \uqd{} approaches} \label{app:paramschoice}

In this section, we study the impact of varying the parameter values for each of the \uqd{} approaches. 
We use the hyperparameters of MAP-Elites given in Appendix \ref{app:hyperparam} for all algorithms.
\uqd{} algorithms require only two additional categories of hyperparameters that are not defined for MAP-Elites: number of samples $N$ and depth $D$.
We replicate \replicationsparams{} times each algorithm on each task for each sampling-size value, for a total of \totalreplicationsparams{} runs this section only.
We display the same pareto graphs as in Sec.~VIII.

\subsection{MAP-Elites-sampling}

MAP-Elites-sampling approaches do not require any depth 
so we only vary the number of samples $N$. We consider the following values: $32, 64, 128, 256$ and display the comparison in Fig.~\ref{fig:MAP-Elites-sampling}. 
The results indicate that the value giving the best performance on average across environments is $N=32$.

\subsection{Deep-Grid}

Deep-Grid does not require any number of samples, all solutions are evaluated once. Thus we only vary the depth of the grid $D$. The original Deep-grid paper \cite{deepgrid} was using $D=50$, we thus compare value similar values. However, as we are working with hardware-accelerated libraries we only consider powers of 2: $32, 64, 128$. We display the comparison in Fig.~\ref{fig:Deep-Grid}. 
The results indicate that the value giving the best performance on average across environments is $D=32$.

\subsection{Deep-Grid-sampling}

Deep-grid-sampling augments Deep-Grid with sampling, it thus requires both a number of samples $N$ and a depth $D$. We choose to stick with the optimal depth value selected for Deep-Grid: $D=32$, and we only vary the number of samples $N$. We consider the values: $8, 32, 64$. 
We display the comparison in Fig.~\ref{fig:Deep-Grid-sampling}. 
The results indicate that the value giving the best performance on average across environments is $N=8$.

\subsection{\archivesampling{}}

\archivesampling{} does not require any number of samples, all solutions are evaluated once. Thus we only vary the depth of the grid $D$.
At each generation of \archivesampling{}, the full content of the grid is reevaluated. For a grid of $N_{cells}$ cells, increasing the depth from $D=2$ to $D=4$ already adds $2 * N_{cells}$ evaluations at each generation. As our comparison is based on sampling-budget, this can totally modify the dynamic of the algorithm, and in extreme scenarios even prevent the algorithm from running. Thus, we choose to focus on small values of $D$. 
We consider the following values: $1, 2, 4$, we consider $D=1$ to show the advantage of having a depth in \archivesampling{} approaches. We display the comparison in Fig.~\ref{fig:Archive-Sampling}. 
The results indicate that the value giving the best performance on average across environments is $D=2$.

\subsection{\eas{}}

\eas{} does not require any number of samples, as it is chosen dynamically based on the content of the grid. Thus we only vary the depth of the grid $D$.
For the exact same reasons mentioned for \archivesampling{}, we choose to focus on small values of $D$ and consider the following values: $1, 2, 4$. Again, we consider $D=1$ to show the advantage of having a depth in \eas{} approaches. We display the comparison in Fig.~\ref{fig:Parallel-Adaptive-Sampling}. 
The results indicate that the value giving the best performance on average across environments is $D=2$.

\begin{figure}[ht]
\centering
\includegraphics[width = 0.85\hsize]{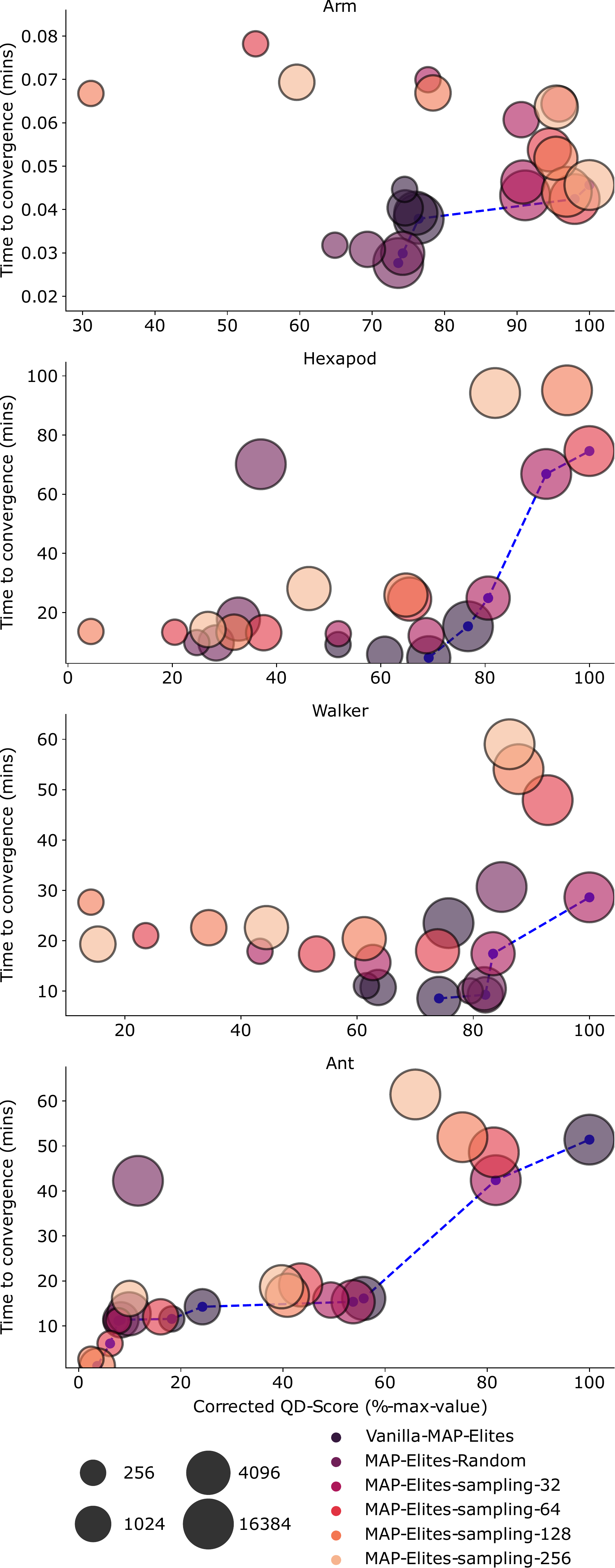}
\caption{
    Pareto front for different parameter values for MAP-Elites-sampling. 
}
\label{fig:MAP-Elites-sampling}
\end{figure}

\begin{figure}[ht]
\centering
\includegraphics[width = 0.85\hsize]{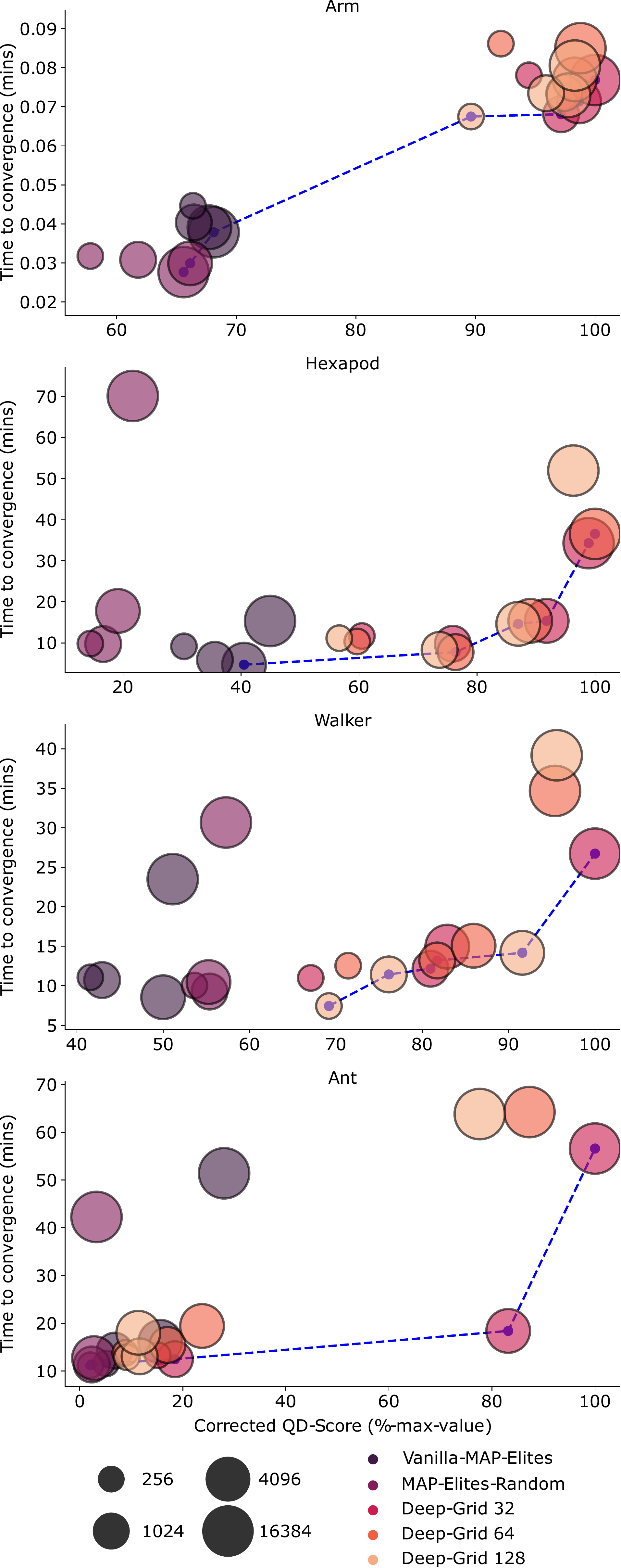}
\caption{
    Pareto front for different parameter values for Deep-Grid. 
}
\label{fig:Deep-Grid}
\end{figure}

\begin{figure}[ht]
\centering
\includegraphics[width = 0.85\hsize]{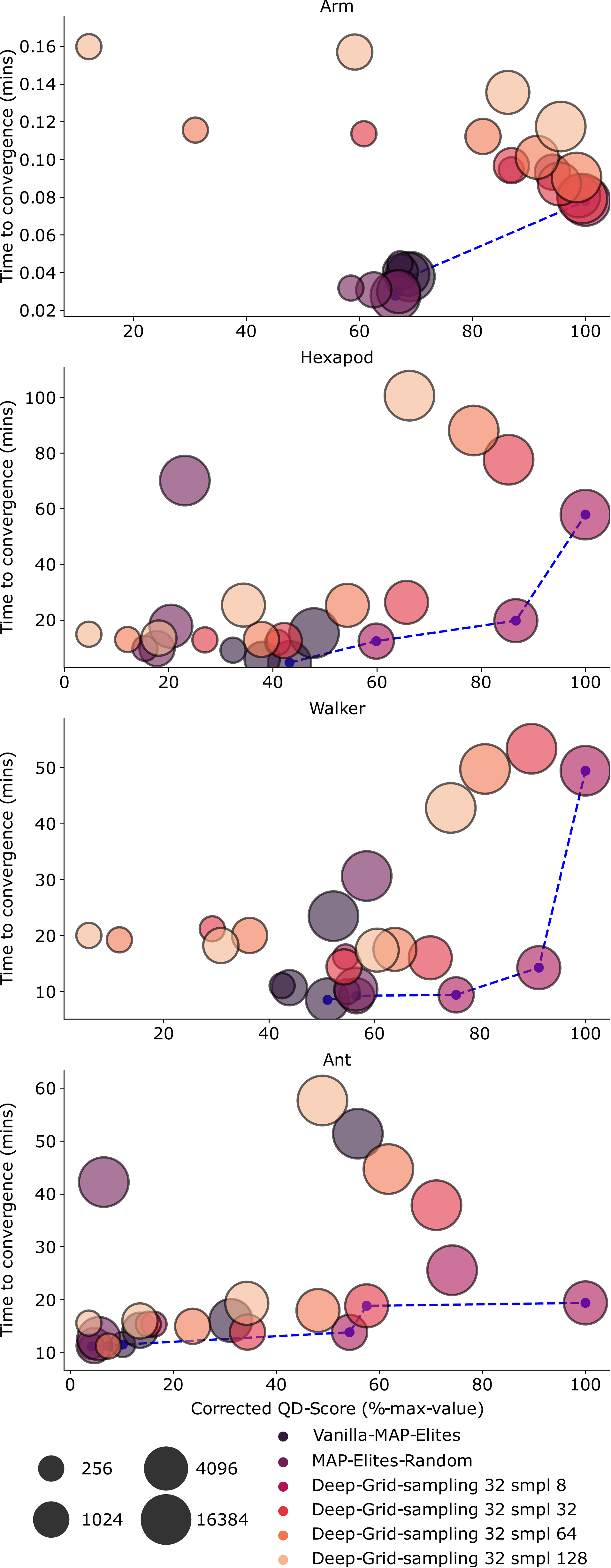}
\caption{
    Pareto front for different parameter values for Deep-Grid-sampling. 
}
\label{fig:Deep-Grid-sampling}
\end{figure}

\begin{figure}[ht]
\centering
\includegraphics[width = 0.85\hsize]{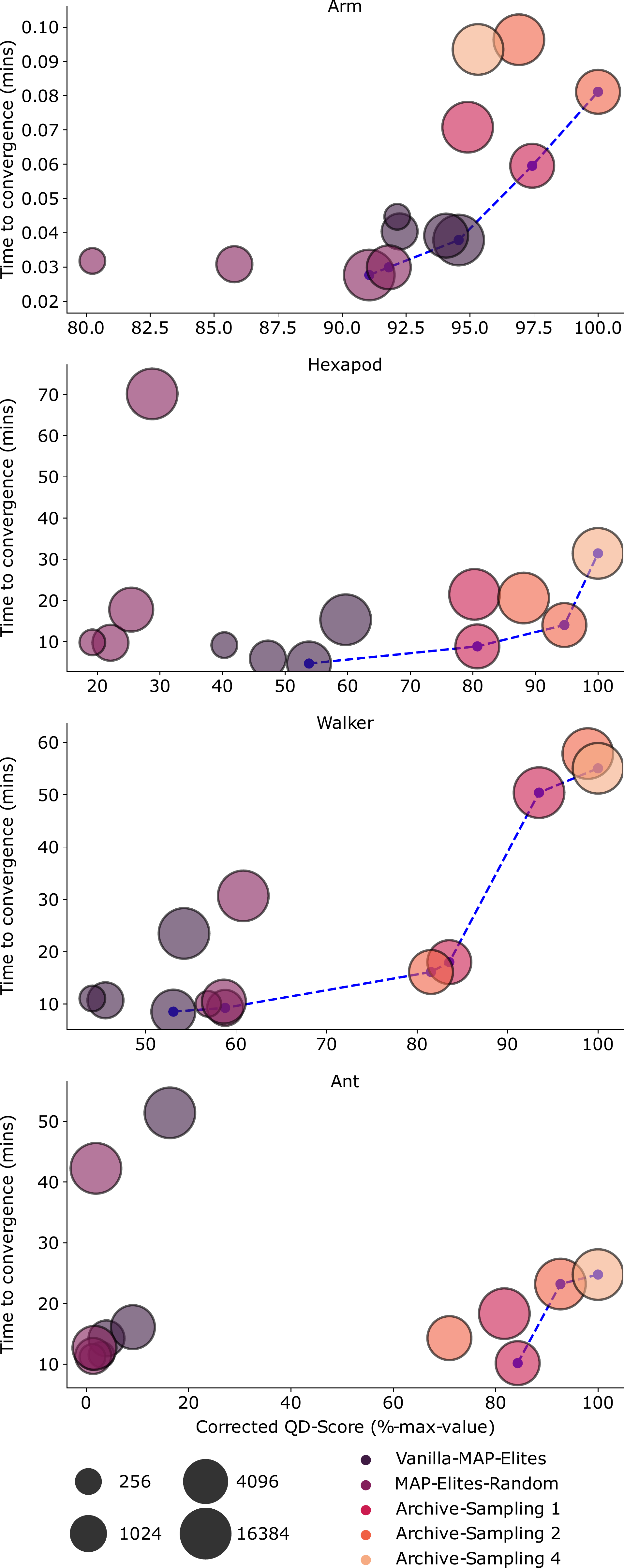}
\caption{
    Pareto front for different parameter values for \archivesampling{}. 
}
\label{fig:Archive-Sampling}
\end{figure}

\begin{figure}[ht]
\centering
\includegraphics[width = 0.85\hsize]{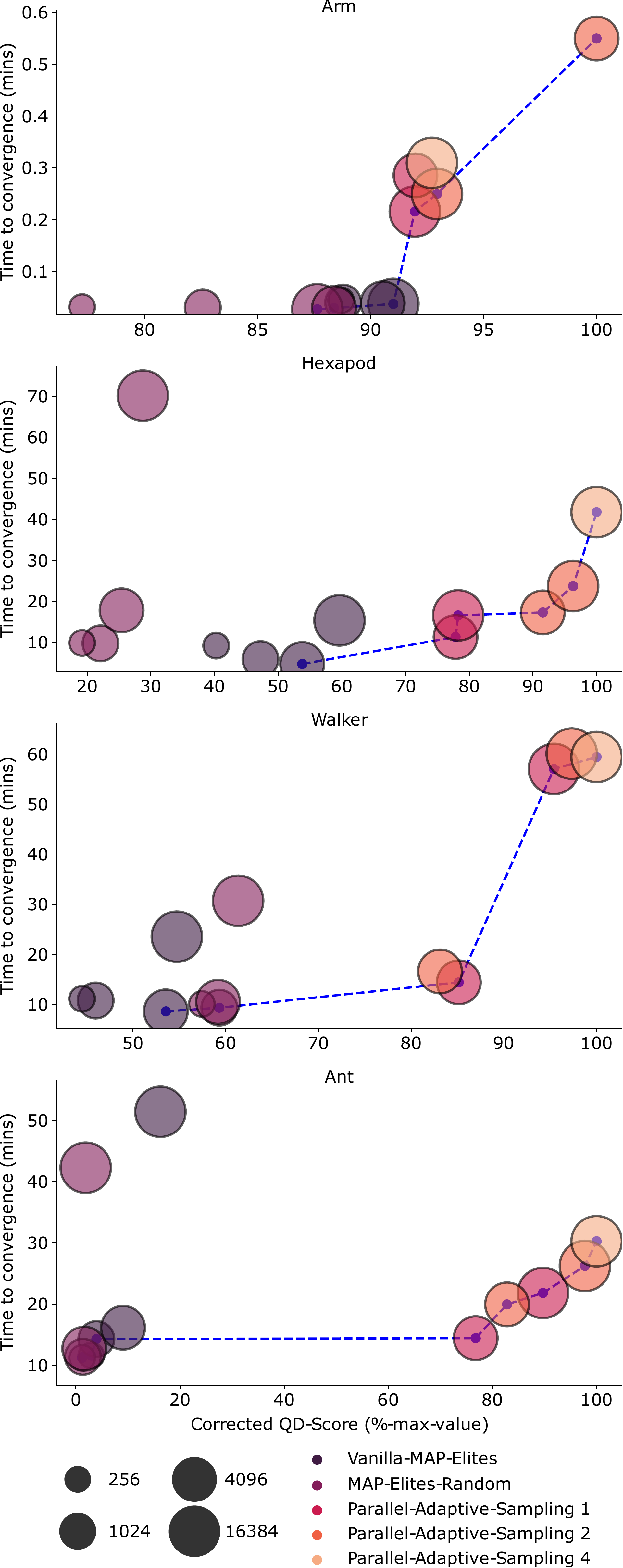}
\caption{
    Pareto front for different parameters for \eas{}. 
}
\label{fig:Parallel-Adaptive-Sampling}
\end{figure}

}

\end{document}